\documentclass[11pt,a4paper]{article}
\usepackage[hyperref]{eacl2021}
\usepackage{times}
\usepackage{latexsym}

\usepackage{amsfonts,amsmath}
\usepackage{array}
\usepackage{graphicx}
\usepackage{subcaption}
\usepackage{multirow}
\usepackage{changepage}
\usepackage{booktabs}

\usepackage{microtype}

\aclfinalcopy 

\title{RuSentEval: Linguistic Source, Encoder Force!}

\author{Vladislav Mikhailov\textsuperscript{1,2},~ Ekaterina Taktasheva\textsuperscript{2},~Elina Sigdel\textsuperscript{2},~Ekaterina Artemova\textsuperscript{2}  \\
  \textsuperscript{1} SberDevices, Sberbank, Moscow, Russia\\
  \textsuperscript{2} HSE University, Moscow, Russia\\
  \tt{Mikhaylov.V.Nikola@sberbank.ru} \\
  {\tt \{etaktasheva, essigdel, elartemova\}{\tt @hse.ru}}
\\}

\date{}

\begin{document}
\maketitle
\begin{abstract}

The success of pre-trained transformer language models has brought a great deal of interest on how these models work, and what they learn about language. However, prior research in the field is mainly devoted to English, and little is known regarding other languages. To this end, we introduce RuSentEval, an enhanced set of 14 probing tasks for Russian, including ones that have not been explored yet. We apply a combination of complementary probing methods to explore the distribution of various linguistic properties in five multilingual transformers for two typologically contrasting languages -- Russian and English. Our results provide intriguing findings that contradict the common understanding of how linguistic knowledge is represented, and demonstrate that some properties are learned in a similar manner despite the language differences.

\end{abstract}

\section{Introduction}
Transformer language models \citep{vaswani2017attention} have achieved state-of-the-art results on a wide range of NLP tasks in multiple languages, demonstrated strong performance in zero-shot cross-lingual transfer \citep{pires2019multilingual}, and even surpassed human solvers in NLU benchmarks such as SuperGLUE \citep{wang2019superglue}. 
The success has stimulated research in how these models work, and what they acquire about language. The majority of the introspection techniques are based on the concept of \emph{probing tasks} \citep{adi2016fine,shi2016does,conneau2018you} which allow analyzing what linguistic properties are encoded in the intermediate representations. A rich variety of tasks has been introduced so far, ranging from token-level and sub-sentence probing \citep{liu2019linguistic,tenney2019you} to sentence-level probing \citep{alt2020probing}. A prominent method to explore the inner workings of the models involves training a lightweight classifier to solve a probing task over features produced by them, and assess their knowledge by the classifier’s performance. Recently, the methods have been greatly extended to latent subclass learning \citep{michael2020asking}, correlation similarity measures \citep{wu2020similarity}, information-theoretic probing \citep{voita2020information}, investigation of individual neurons \citep{durrani2020analyzing,suau2020finding}, and many more.

Despite growing interest in the field, English remains the focal point of prior research \citep{belinkov2019analysis,rogers2020primer} leaving other languages understudied. To this end, several monolingual and cross-lingual probing suites have been assembled (see Section \ref{sec:related_work}), with a few of them following SentEval toolkit \citep{conneau2018you,conneau2018senteval}. However, most of them directly apply an English-oriented method which is not guaranteed to be universal across languages, or use the Universal Dependencies (UD) Treebanks \citep{nivre2016universal} which tend to be inconsistent \citep{alzetta2017dangerous,de2017assessing,droganova2018data}.

This work proposes \textbf{RuSentEval}, a probing suite for evaluation of sentence embeddings for the Russian language. We adapted the method for English \citep{conneau2018you} to complement the peculiarities of Russian. In contrast to closely related datasets \citep{ravishankar2019probing,eger2020probe}, RuSentEval is fully guided by linguistic expertise, relies on annotations obtained with the current state-of-the-art model for Russian morphosyntactic analysis \citep{anastasyev2020}, and includes tasks that have not been explored yet.

The contributions are summarized as three-fold. First, we present an enhanced set of 14 probing tasks for Russian, organized by the type of linguistic properties. Second, we carry out a series of probing experiments on two typologically different languages: English, which is an analytic Germanic language, and Russian, which is a fusional Slavic one. Keeping in mind that the English \citep{conneau2018you} and Russian datasets are combined from different annotation schemas, we introspect five multilingual transformer-based encoders, including their distilled versions. We apply several probing methods to conduct the analysis from different perspectives, and support the findings with statistical significance. Besides, we establish count-based and neural-based baselines for the tasks. Finally, RuSentEval is publicly available\footnote{\url{https://github.com/RussianNLP/rusenteval}}, and we hope it will be used for evaluation and interpretation of language models and sentence embeddings for Russian.

\section{Related Work}
\label{sec:related_work}
Introspection of pre-trained language models for languages other than English, specifically Russian, is usually conducted in the cross-lingual setting. The primary goal of such experiments is to explore how particular linguistic properties are distributed in a given collection of multilingual embedding and language models. LINSPECTOR \citep{csahin2020linspector} is one of the first probing suites that covers a wide range of linguistic phenomena in 24 languages. The benchmark uses UniMorph 2.0 \citep{kirov2018unimorph} to design the tasks since UD do not provide a sufficient amount of data for the considered languages. Despite this, UD Treebanks have become the main source for collection of multilingual probing tasks, limiting the scope to morphology and syntax \citep{krasnowska2019empirical}. Other sources for the assembly include multilingual datasets created by means of machine translation, such as XNLI \citep{conneau2018xnli}, or datasets that labelled with similar annotation schemes for named entity recognition (NER) and semantic role labelling (SRL) tasks \citep{csahin2020linspector}.

A few prior works \citep{ravishankar2019probing,eger2020probe} that follow SentEval toolkit \citep{conneau2018senteval} directly apply the method designed for English to multiple typologically diverse languages, which raises doubts if such strategy is universal across languages that exhibit unique peculiarities, e.g. a free word order and rich inflectional morphology. This can lead to low quality of the datasets and unreliable experimental results, particularly for Russian. Consider a few examples for (\textbf{BShift}) task in Russian: \textit{\textbf{Fedra zatem} povesilas', a Tesey uznal pravdu.} ``\textbf{Phaedra later} hanged herself, and Theseus unraveled the truth.'' \citep{ravishankar2019probing}, and \textit{Shestoe – \textbf{zanimatsya nado} svoim obrazovaniem} ``The sixth point is that you \textbf{to take care need} of your education'' \citep{eger2020probe}. The sentences are labelled as positive, meaning that they exhibit incorrect word order. While the word order changes can lead to the syntax perturbations for English, both sentences are still acceptable in terms of syntax for Russian. Moreover, the dataset sizes tend to be inconsistent across languages due to using UD Treebanks which makes it difficult to compare the results \citep{eger2020probe}.

Another line of research includes probing machine translation models \citep{marevcek2020multilingual} over multiple languages, and probing for cross-lingual similarity by utilizing paired sentences in mutually intelligible languages \citep{choenni2020cross, choenni2020does}. Last but not least, such benchmarks as XGLUE \cite{liang2020xglue} and XTREME \cite{hu2020xtreme} allow evaluating the current state of language transferring methods.

\section{Probing Tasks}
\label{sec:probing_tasks}

\paragraph{Data} The sentences for our probing tasks were extracted from the following publicly available resources: Russian Wikipedia articles\footnote{\url{https://dumps.wikimedia.org/ruwiki/latest/}} and news corpora such as Lenta.ru\footnote{\url{https://github.com/yutkin/Lenta.Ru-News-Dataset}} and the news segment of Taiga corpus \citep{shavrina2017methodology}. We used rusenttokenize library\footnote{\url{https://pypi.org/project/rusenttokenize}}, a rule-based
sentence segmenter for Russian, to split texts into sentences. Each sentence was tokenized with spaCy Russian Tokenizer\footnote{\url{https://github.com/aatimofeev/spacy_russian_tokenizer}}. The sentences were filtered by the 5-to-25 token range and annotated with the current state-of-the-art model for joint morphosyntactic analysis in Russian \citep{anastasyev2020}. Besides, we performed two additional preprocessing steps. (1) We computed the IPM frequency of each sentence using the New Frequency Vocabulary of Russian Words \cite{lyashevskaya2009frequency} to control the word frequency distribution. The IPM values of each token lemma in the sentence (if present in the vocabulary) were averaged over a total number of token lemmas. The sentences with the IPM frequency of lower than 0.9 were filtered out. This allows discarding the majority of sentences that contain rare words, acronyms, abbreviations, or loanwords. (2) Syncretism is peculiar to fusional languages, that is, when a word can belong to multiple part-of-speech tags, or express multiple ambiguous morphosyntactic features \citep{baerman2007syncretism}. Following \citep{csahin2020linspector}, we removed sentences in the semantic tasks (described below) where the target word has multiple part-of-speech tags. This step allows simplifying the probe interpretation and ensuring a fairer experimental setup in terms of the language comparison.

The total number of the annotated sentences after filtering is 3.6 mln, and they are publicly available. Each task consists of a 100k-sentence training set and 10k-sentence validation and test sets. There is no sentence overlap across the splits, and all sets are balanced by the number of instances per target class.

\vspace{0.5em}\noindent \textbf{Surface properties} tasks test if it is possible to recover information about surface properties from the contextualized representations. (\textbf{SentLen}) is a 6-way classification task aimed to predict a number of tokens given a sentence representation. Similar to \cite{adi2016fine,conneau2018you}, we grouped sentences into 6 equal-width bins by length. The \emph{word content} (\textbf{WC}) task tests if the information on the original words in a sentence can be inferred from its representation. We selected 1k lemmas from the source corpus vocabulary within the 1.5k-3k rank range when sorted by frequency, and sampled equal numbers of sentences that contain only one of these lemmas. The task is treated as a 1k-way classification that requires knowledge about lexical items and their inflectional paradigms.

\vspace{0.5em}\noindent \textbf{Syntactic properties} is a group of tasks that probe the encoder representations for syntactic properties. In the (\textbf{ConjType}) task, the sentences must be classified in terms of the type of connection between complex clauses. The objective of the classifier is to tell whether a sentence involves coordination or subordination. 

(\textbf{ImpersonalSent}) is a binary classification task that aims to define if there is a lack of a grammatical subject in the main clause of a sentence. It is usually expressed by a singular third-person, reflexive, singular neuter, or invariable predicate form (\textit{smerkaetsya} ``it is getting dark''; \textit{zharko} ``it is hot''; \textit{pora idti} ``it is time to go''), an adverbial predicate phrase (\textit{bylo sovershenno tikho} ``it was absolutely quiet''), and intransitive verbs typically combined with a noun phrase in the instrumental (\textit{zapahlo rozami} ``it smells roses'').

(\textbf{TreeDepth}) task tests whether the encoder representations store the information on the hierarchical and syntactic structure of sentences. Specifically, the goal is to probe for knowledge about the depth of the longest path from the root node to any leaf in the syntax tree. Similar to \cite{conneau2018you}, we obtained sentences where the tree depth and the sentence length are de-correlated. The tree depth values range from 5 to 9 which makes (\textbf{TreeDepth}) a 5-way classification task.

(\textbf{Gapping}) task deals with the detection of syntactic gapping that occurs in coordinated structures and elides a repeated predicate, typically from the second clause. We used data provided in the Shared Task on Automatic Gapping Resolution for Russian, or AGRR-2019 \cite{ponomareva2019agrr}. For instance, the sentence \textit{Odin imel silu solntsa, drugoy – luny.} ``One had the power of the Sun, the other \textit{\textbf{(had the power of)}} the Moon.'' comprises an omission of a repeating predicate in the non-initial clause with its semantics remaining expressed.

The \emph{N-gram shift} task (\textbf{NShift}) is analogous to SentEval's (\textbf{Bshift}) that tests the encoder's sensitivity to incorrect word order. As opposed to English, only specific cases of word inversion in Russian lead to syntax perturbation. We, therefore, perturbed N-grams that correspond to a set of pre-defined morphosyntactic patterns. We used TF-IDF method from scikit-learn library \cite{pedregosa2011scikit} to build an N-gram feature matrix that was further applied for the word order perturbation. For instance, we reversed adjacent words in prepositional phrases, numeral phrases, compound noun phrases, etc. Below is an example where the head of the prepositional phrase \textit{v schkolu} 'to school' is inverted with the dependent noun:
\begin{align*}
\text{Segodnya on ne poshel \textbf{shkolu v}.}\\
\text{``He did not go \textbf{school to} today.'}
\end{align*}

\vspace{0.5em}\noindent \textbf{Semantic properties} tasks rely on both syntactic and semantic structure of a sentence to recover a higher-level property. (\textbf{SubjNumber}) and (\textbf{SubjGender}) probe for the number and gender features of the subject in the main clause. Similarly, (\textbf{ObjNumber}) and (\textbf{ObjGender}) focus on the number and gender of direct object in the main clause.  
In the following tasks, the aim is to probe for the morphosyntactic features of the predicate or the head of a predicative construction of the main clause: \emph{predicate voice} (\textbf{PV}), \emph{predicate aspect} (\textbf{PA}), and \emph{predicate tense} (\textbf{PT}). The latter is analogous to SentEval's (\textbf{Tense}) task.

The semantic tasks test if the contextualized representations not only capture the morphosyntactic features but also encode higher-level, structural and syntactic-semantic information (namely, the syntax tree hierarchy and the actant structure of a predicate). Note that the boundary between the surface, syntactic and semantic tasks is relatively blurred.

\section{Experimental Setup}
\label{sec:experimental_setup}

\begin{table*}[t!]
\centering
\begin{tabular}{c|c|c|c|c|c|c}
\toprule
\textbf{Probing Task} & \textbf{Language} & \textbf{M-BERT} & \textbf{LABSE} & \textbf{XLM-R} &  \textbf{MiniLM} & \textbf{M-BART} \\
\midrule
\textbf{Nshift} & 
    \begin{tabular}{@{}c@{}}\textbf{Ru} \\ \textbf{En}\end{tabular}
  & \begin{tabular}{@{}c@{}} 84.8 [8] \\ 81.8 [10] \end{tabular} & \begin{tabular}{@{}c@{}}  82.6 [5] \\ 84.4 [5] \end{tabular} & \begin{tabular}{@{}c@{}} \colorbox{lightgray}{\textbf{86.9 [9]}} \\ \colorbox{lightgray}{\textbf{85.7 [10]}} \end{tabular} & \begin{tabular}{@{}c@{}} 80.5 [9] \\ 79.3 [8] \end{tabular} & \begin{tabular}{@{}c@{}} 78.6 [12] \\ 83.8 [12] \end{tabular}
  \\ \midrule

\textbf{ObjNumber} & 
    \begin{tabular}{@{}c@{}}\textbf{Ru} \\ \textbf{En}\end{tabular}
  & \begin{tabular}{@{}c@{}} 82.8 [6] \\  \colorbox{lightgray}{\textbf{86.2 [6]}} \end{tabular} & \begin{tabular}{@{}c@{}} 82.5 [2] \\ 85.4 [3] \end{tabular} & \begin{tabular}{@{}c@{}} \colorbox{lightgray}{\textbf{83.7 [10]}} \\ 86.0 [8] \end{tabular} & \begin{tabular}{@{}c@{}} 77.8 [10] \\ 85.2 [6] \end{tabular} & \begin{tabular}{@{}c@{}} 81.5 [7] \\ 85.9 [9] \end{tabular}
  \\ \midrule

\textbf{SentLen} & 
    \begin{tabular}{@{}c@{}}\textbf{Ru} \\ \textbf{En}\end{tabular}
  & \begin{tabular}{@{}c@{}} 91.3 [2] \\ 96.3 [2] \end{tabular} & \begin{tabular}{@{}c@{}} 93.3 [1] \\ 96.6 [1] \end{tabular} & \begin{tabular}{@{}c@{}} 94.5 [2] \\ 95.8 [2] \end{tabular} & \begin{tabular}{@{}c@{}} 94.1 [2] \\ 96.1 [3] \end{tabular} & \begin{tabular}{@{}c@{}} \colorbox{lightgray}{\textbf{96.2 [4]}} \\ \colorbox{lightgray}{\textbf{97.3 [3]}} \end{tabular}
  \\ \midrule
  
  \textbf{SubjNumber} & 
    \begin{tabular}{@{}c@{}}\textbf{Ru} \\ \textbf{En}\end{tabular}
  & \begin{tabular}{@{}c@{}} 90.5 [7] \\ 87.8 [7] \end{tabular} & \begin{tabular}{@{}c@{}} 92.9 [3] \\ \colorbox{lightgray}{\textbf{90.7 [12]}} \end{tabular} & \begin{tabular}{@{}c@{}} \colorbox{lightgray}{\textbf{94.9 [11]}} \\ 86.9 [10] \end{tabular} & \begin{tabular}{@{}c@{}} 94.2 [12] \\ 85.6 [6] \end{tabular} & \begin{tabular}{@{}c@{}} 93.1 [10] \\ 87.3 [9] \end{tabular}
  \\ \midrule
  
  \textbf{Tense} & 
    \begin{tabular}{@{}c@{}}\textbf{Ru} \\ \textbf{En}\end{tabular}
  & \begin{tabular}{@{}c@{}} 99.5 [8] \\ 88.9 [8] \end{tabular} & \begin{tabular}{@{}c@{}} \colorbox{lightgray}{\textbf{99.8 [5]}} \\ 88.8 [6] \end{tabular} & \begin{tabular}{@{}c@{}} \colorbox{lightgray}{\textbf{99.8 [5]}} \\ 88.8 [9] \end{tabular} & \begin{tabular}{@{}c@{}} 98.2 [7] \\ 87.3 [5] \end{tabular} & \begin{tabular}{@{}c@{}} 99.6 [7] \\ \colorbox{lightgray}{\textbf{89.1 [9]}} \end{tabular}
  \\ \midrule
  
  \textbf{TreeDepth} & 
    \begin{tabular}{@{}c@{}}\textbf{Ru} \\ \textbf{En}\end{tabular}
  & \begin{tabular}{@{}c@{}} 44.7 [6] \\ 41.2 [5] \end{tabular} & \begin{tabular}{@{}c@{}} 46.1 [4] \\ \colorbox{lightgray}{\textbf{42.7 [5]}} \end{tabular} & \begin{tabular}{@{}c@{}} \colorbox{lightgray}{\textbf{46.5 [5]}} \\ 41.8 [7] \end{tabular} & \begin{tabular}{@{}c@{}} 44.8 [7] \\ 40.9 [7] \end{tabular} & \begin{tabular}{@{}c@{}} 45.8 [11] \\ 41.2 [12] \end{tabular}
  \\ \midrule
  
  \textbf{WC} & 
    \begin{tabular}{@{}c@{}}\textbf{Ru} \\ \textbf{En}\end{tabular}
  & \begin{tabular}{@{}c@{}} 84.8 [2] \\ 92.6 [1] \end{tabular} & \begin{tabular}{@{}c@{}} 85.8 [1] \\ 93.7 [1] \end{tabular} & \begin{tabular}{@{}c@{}} 82.6 [1] \\ 89.8 [1] \end{tabular} & \begin{tabular}{@{}c@{}} 72.8 [1] \\ 82.3 [1] \end{tabular} & \begin{tabular}{@{}c@{}}  \colorbox{lightgray}{\textbf{88.0 [1]}} \\ \colorbox{lightgray}{\textbf{93.8 [1]}} \end{tabular}
  \\ 
  \bottomrule
  
\end{tabular}
\caption{Results of Logistic Regression classifier for each encoder over the shared English and Russian tasks. Languages: \textbf{Ru}=Russian, \textbf{En}=English.}
\label{tab:intersection_results_logreg}
\end{table*}

\subsection{Encoders}
We run the experiments on the following 12-layer multilingual transformer encoders released by HuggingFace \citep{wolf2019huggingface}:

\vspace{0.5em}\noindent \textbf{M-BERT} \citep{devlin2018bert} is trained on masked language modeling (MLM) and next sentence prediction tasks, over concatenated monolingual Wikipedia corpora in 104 languages.

\vspace{0.5em}\noindent \textbf{XLM-R} \citep{conneau2019unsupervised} is trained on 'dynamic' MLM task, over filtered CommonCrawl data in 100 languages \citep{wenzek2019ccnet}.

\vspace{0.5em}\noindent \textbf{MiniLM} \citep{wang2020minilm} is a distilled transformer of BERT architecture, but uses the XLM-RoBERTa tokenizer.

\vspace{0.5em}\noindent \textbf{LABSE} \citep{feng2020language} employs a dual-encoder architecture that combines MLM and translation language modeling  \citep{conneau2019cross}.

\vspace{0.5em}\noindent \textbf{M-BART} \citep{liu2020multilingual} is a sequence-to-sequence transformer model with a BERT encoder, and an autoregressive GPT-2 decoder \citep{radford2019language}. We used only the encoder in the experiments.

\subsection{Methods}
\label{subsec:probe_methods}
\paragraph{Probing Classifiers}
We trained linear and non-linear classifiers over intermediate representations produced by the encoders\footnote{We used mean-pooled sentence representations to train the classifiers.}, using categorical cross-entropy loss, and Adam optimizer \citep{kingma2014adam}. For the non-linear classifier (MLP), we used the Sigmoid activation function, the number of hidden states of 250, and the dropout rate of 0.2. Training is run over 5 iterations with the L2-regularization parameter $\in [0.1, ..., 1\emph{e}^{-5}]$ tuned on the validation set, and the best classifier selected. The performance is evaluated by accuracy score.

\paragraph{Individual Neuron Analysis} Neuron-level introspection technique \citep{durrani2020analyzing} allows identifying \emph{top neurons} that contribute most to a probing task and observe how these neurons are distributed across layers of the encoder. Similarly, we trained a linear probing classifier over concatenated sentence representations and used the weights to measure the importance of each neuron. The classifier is trained using Elastic-net regularization with L1 and L2 $\lambda$'s $\in [0.1, \ldots, 1\emph{e}^{-5}]$ tuned on the validation set. Refer to \citep{dalvi2019one,durrani2020analyzing} for more details. 

\paragraph{Correlation Analysis}
Correlation-based analysis techniques posed in \citep{wu2020similarity} allow measuring similarity of the encoder intermediate representations without any linguistic annotation. We apply neuron-level (\texttt{maxcorr}) and representation-level (\texttt{lincka}) similarity measures to investigate the encoders. \texttt{maxcorr} identifies pairs of neurons of the maximum correlation from two different layers. It is high when two layers have pairs of neurons with similar behavior. \texttt{lincka} provides a comparison of representations from different layers in a given collection of models. Two layers are assigned a high similarity if their representations behave similarly over all the neurons.

\subsection{Baselines}
We established a number of count-based and non-contextualized baseline features to train the probing classifiers as outlined in Section \ref{subsec:probe_methods}. We used N-gram range $\in [1,4]$ and top-150k features in the vocabularies for each count-based baseline. The count-based features include \textbf{TF-IDF over character N-grams}, \textbf{TF-IDF over BPE tokens} \citep{sennrich2015neural}, and \textbf{TF-IDF over SentencePiece tokens} \citep{kudo2018sentencepiece}. We applied multilingual BertTokenizer and XLMRobertaTokenizer by HuggingFace to segment sentences into BPE and SentencePiece tokens. The non-contextualized features are mean-pooled monolingual \textbf{fastText} sentence embeddings \cite{bojanowski2017enriching}. We used monolingual fastText models for English\footnote{\url{https://fasttext.cc/docs/en/crawl-vectors.html}} and Russian. The latter was trained over joint Russian Wikipedia and Lenta.ru news, and released by DeepPavlov \citep{burtsev2018deeppavlov}.

\section{Results}

\label{sec:results}
How is the linguistic knowledge of two contrasting languages distributed in pre-trained multilingual encoders? This section describes the results for shared English and Russian probing tasks, covering surface properties (\textbf{SentLen}, \textbf{WC}), syntax (\textbf{TreeDepth}, \textbf{NShift}), and semantics (\textbf{ObjNumber}, \textbf{SubjNumber}, and \textbf{Tense}). We also report the results for the remaining tasks and the baselines in Appendix \ref{probe_res}.

\begin{figure*}[t!]
  \centering
  \includegraphics[width=.92\textwidth]{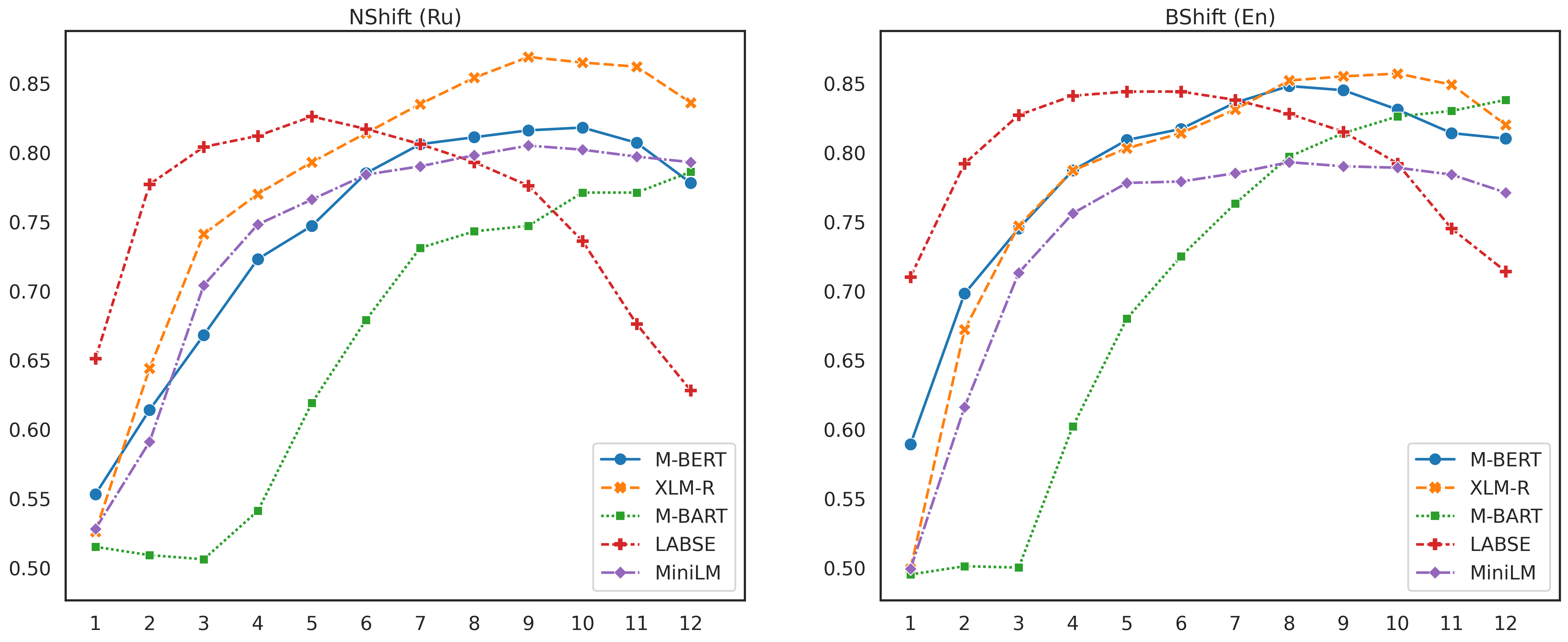}
  \caption{The probing results for each encoder on \textbf{NShift} (\textbf{Ru}, left) and \textbf{Bshift} (\textbf{En}, right) tasks. X-axis=Layer index number, Y-axis=Accuracy score.}
  \label{fig:ru_en_nshift_logreg}
\end{figure*}

\subsection{Layer-wise Supervised Probing}
\label{subsection:sup_probe}

Table \ref{tab:intersection_results_logreg} presents the results of the linear classifier performance over the shared tasks. For the sake of space, we omit the results of the non-linear classifier and provide them in Appendix \ref{probe_res}. The best score for each task in each language is highlighted in grey, and the index number of the layer achieving the score is enclosed in square brackets. We observe that the surface and syntax tasks show similar trends for both languages, while a few semantic tasks reveal some differences.
The overall pattern for the surface-level tasks (\textbf{SentLen}, \textbf{WC}) is that the probing curves\footnote{We refer to \emph{probing curves} as to the performance trajectory of a probing classifier} for both languages tend to be decaying after reaching the peak at the lower layers $[1-4]$. The exception is provided by \textbf{M-BART} which keeps the surface properties across the majority of layers. The baselines mostly perform on par with one another, with \textbf{fastText} and \textbf{TF-IDF Char} receiving the best score (see Appendix \ref{probe_res}).

(\textbf{Nshift}, \textbf{TreeDepth}) tasks demonstrate interesting distinctive features of \textbf{LABSE} and \textbf{M-BART} encoders with respect to the syntactic properties. Figure \ref{fig:ru_en_nshift_logreg} depicts the probing curves over (\textbf{Nshift}) and (\textbf{Bshift}) tasks. First, \textbf{LABSE} is at most sensitive to the incorrect word order at the lower-to-middle layers $[2-8]$, as opposed to other encoders which distribute the information at the middle-to-higher layers $[8-12]$. Second, \textbf{M-BART} shows confident knowledge in \textbf{TreeDepth} task for both languages at the middle-to-higher layers $[7-12]$ and reaches the peak at layer 12, in contrast to the rest of the encoders which generally distribute the property at the middle layers $[5-7]$ (see Appendix \ref{probe_res}). The baselines performance varies from achieving a low score (\textbf{TreeDepth}) to being close to random choice (\textbf{Nshift}).

The probing curves for (\textbf{PT}) and (\textbf{Tense}) tasks illustrate that the models encode the property in a very similar fashion, achieving the peak score at the middle layers $[4, 5]$, and flattening the curves until the output layer (see Appendix \ref{probe_res}). In contrast, the behavior of the encoders on (\textbf{SubjNumber}) and (\textbf{ObjNumber}) is slightly different. The number of the subject for English is predominantly distributed at the middle-to-higher levels $[5-12]$, while for Russian it is either at a steady pace from the very first layer (\textbf{M-BART}, \textbf{M-BERT}), or decaying at the lower and middle layers $[3-5]$, and further increasing close to the higher levels $[8-12]$ (\textbf{LABSE}, \textbf{MiniLM}). Other differences are found in the results for (\textbf{ObjNumber}) task. Specifically, the number of direct object for English is best inferred at the lower layers of \textbf{LABSE} $[2, 3]$, as compared to the middle-to-higher layers of other encoders $[6-9]$. Despite this, the property is similarly distributed across the layers of the models. In the same manner as for English, \textbf{LABSE} concentrates the knowledge for the Russian task in layer 2 but decays once reaching the peak. However, the other models exhibit a distinct behavior as well, with the property best encoded at the middle layers $[6, 7]$ (\textbf{M-BERT}, \textbf{M-BART}), or at layer 10 (\textbf{XLM-R}, \textbf{MiniLM}). Notably, the baselines receive a strong performance over the majority of the semantic tasks for both languages, meaning that they can be solved using sub-word features that can capture lexical, or morphosyntactic information. 

\vspace{0.5em}\noindent \textbf{Bootstrap} We estimate statistical significance of the supervised probing by means of layer-wise bootstrap procedure \citep{berg2012empirical}.  Typically, we observe that when the probing curve rises, the layer-wise difference is statistically significant, while at the peak of the probing curve it turns insignificant. We can treat these observations in the following way: starting from the peak of the probing curve, the models acquire the knowledge needed for the task, and do not generally lose it in the higher layers.

\subsection{Neuron-level Analysis}
\label{subsec:neuron_analysis}
\begin{figure*}
  \centering
  \includegraphics[width=\textwidth]{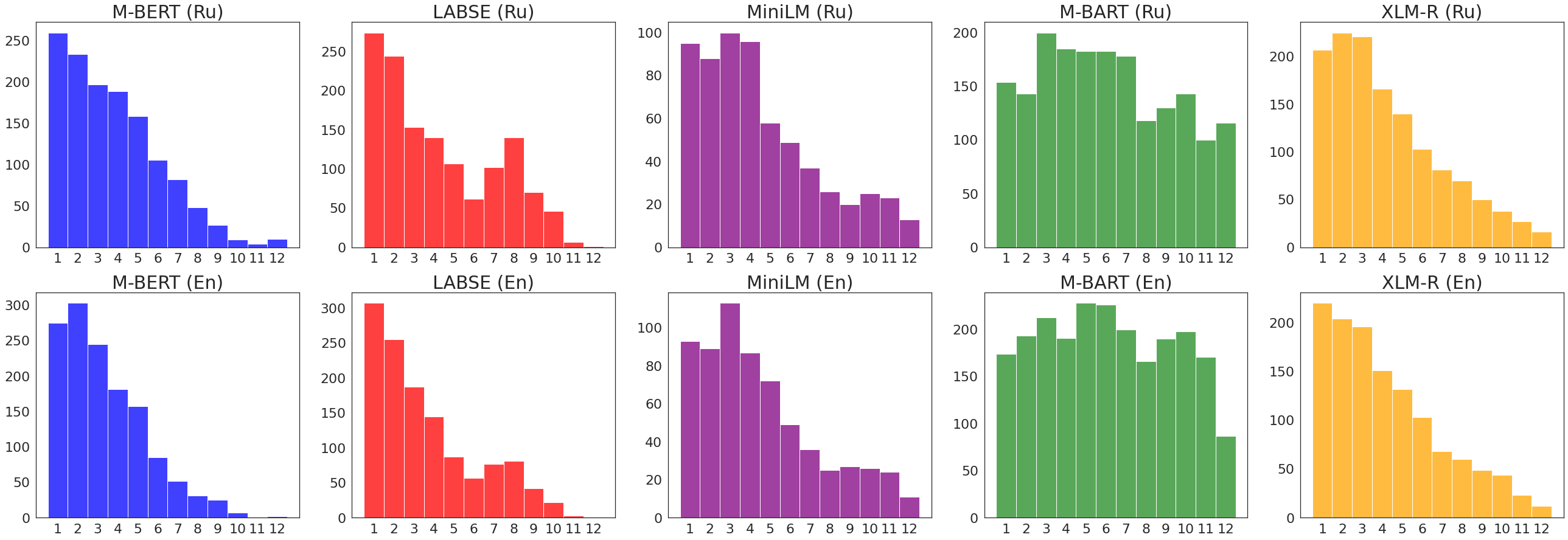}
  \caption{The distribution of top neurons over \textbf{SentLen} tasks for both languages: \textbf{Ru}=Russian, \textbf{En}=English. X-axis=Layer index number, Y-axis=Number of neurons selected from the layer.}
  \label{fig:neuron_sent_len}
\end{figure*}

In contrast to layer-wise supervised probing (Section \ref{subsection:sup_probe}) which introspects each layer independently, individual neuron analysis allows exploring the distribution of top neurons selected from \emph{the entire encoder}. This provides an alternative perspective to which layers contribute predominantly towards the probing tasks. Note that we trained distinct classifiers regularized with Elastic-net~\cite{zou2005regularization} to estimate the importance of the neurons\footnote{We selected top-20\% neurons from each encoder using neuron ranking algorithm described in \citep{durrani2020analyzing}}. Figure \ref{fig:neuron_sent_len} presents the results for (\textbf{SentLen}) tasks in Russian and English. Different from other models which predominantly capture the property by neurons at the lower layers $[1-4]$, \textbf{M-BART} distributes the information across all the layers. Let us analyze the results with respect to some syntactic and semantic levels. A similar behavior of the encoders by language is observed over (\textbf{NShift}) and (\textbf{TreeDepth}) tasks (see Appendix \ref{appendix:neuron_analysis}). \textbf{M-BERT}, \textbf{XLM-R}, and \textbf{MiniLM} generally capture the sensitivity to illegal word order by neurons at the middle and higher layers, while it is contributed by fewer layers of \textbf{LABSE} (\textbf{Ru}: $[2-4]$; \textbf{En}: $[4-7]$), and \textbf{M-BART} which surprisingly stores the property at layer 12 for each language. Another interesting finding is that the depth of the syntax tree is typically spread across most of the layers of each encoder. An exception to this pattern is \textbf{M-BART} which locates the knowledge at the middle-to-higher layers $[7-11]$.

The neuron distributions for (\textbf{ObjNumber}) task are akin by language and slightly different by encoder, showing that the number of the direct object is learned by at least 5-7 layers of different encoder depth. At the same time, the properties for (\textbf{SubjNumber}) and (\textbf{PT}, \textbf{Tense}) tasks are captured differently by the encoders. \textbf{M-BERT}, \textbf{XLM-R} and \textbf{MiniLM} reveal similar behavior by task, whereas \textbf{LABSE} specializes the knowledge at either the lower layers (\textbf{Ru, SubjNumber}; and \textbf{En, Tense}: $[1-4]$), or the higher ones (\textbf{En, SubjNumber}; and \textbf{Ru, PT}: $[10-12]$). On the other hand, the most contributing neurons of \textbf{M-BART} are predominantly spread across all the layers.

\subsection{Correlation Analysis}
\label{subsec:corr_methods}
\begin{figure*}[t!]
    \centering
    \begin{subfigure}[b]{0.49\linewidth}
    \centering
    \includegraphics[width=\linewidth]{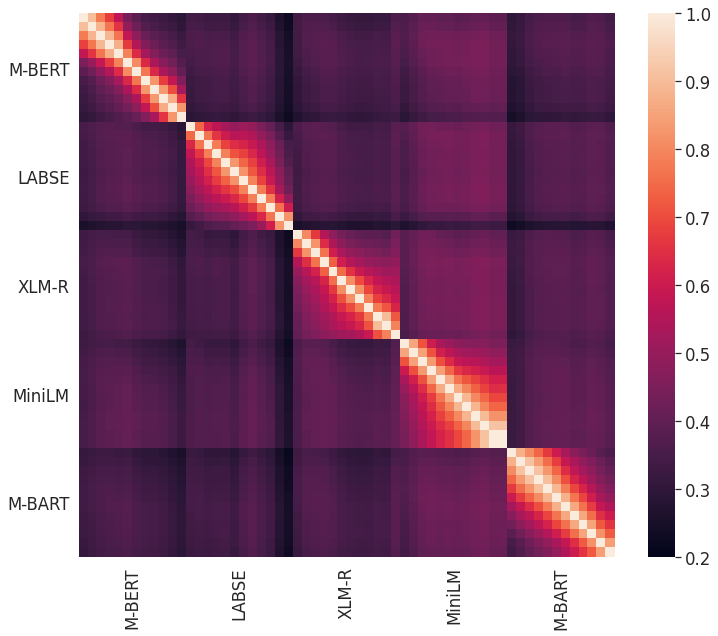}
    \caption{\texttt{maxcorr}}
    \label{fig:heatmap-maxcorr}
    \end{subfigure}
    \begin{subfigure}[b]{0.49\linewidth}
    \centering
    \includegraphics[width=\linewidth]{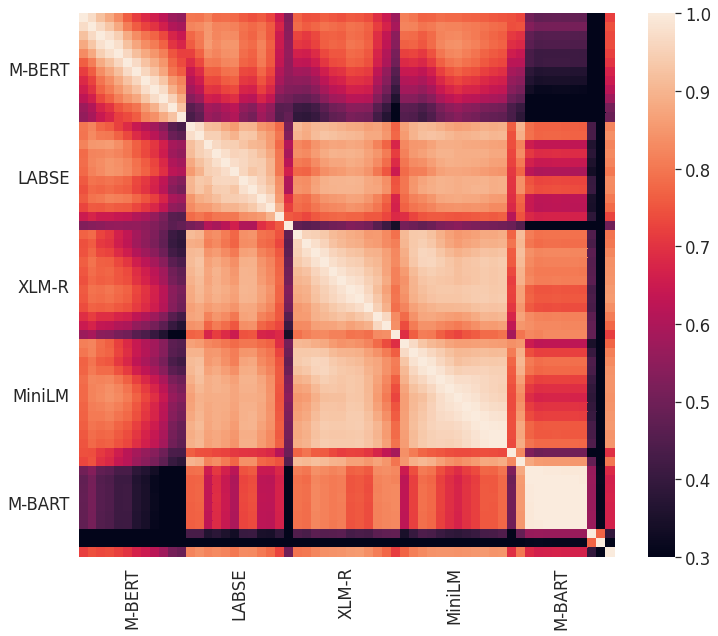}
    \caption{\texttt{lincka}}
    \label{fig:heatmap-lincka}
    \end{subfigure}    
    \caption{Similarity heatmaps of layers in the encoders under neuron-level (\texttt{maxcorr})
    and representation-level (\texttt{lincka}) measures.
    }
    \label{fig:heatmaps}
\end{figure*}

For analyzing the encoders by means of the correlation-based techniques, we used 1k stratified sentences from each test set of the shared tasks. We obtained the sentence representations and computed the measures applying the publicly-available code \citep{wu2020similarity}. Figure \ref{fig:heatmaps} shows heatmaps of similarities between layers of the encoders under neuron-level and representation-level measures for English. Notably, the heatmaps are very alike to the ones for Russian, which we enclose in Appendix \ref{appendix:corr_methods}. \texttt{maxcorr} (Figure \ref{fig:heatmap-maxcorr}) demonstrates that different layers of a single encoder have similar individual neurons, but the inter-encoder neuron similarities are greatly low. On the contrary, \texttt{lincka} (Figure \ref{fig:heatmap-lincka}) induces considerably high similarities across the encoders, meaning that they produce similar sentence representations. However, \textbf{M-BERT} and \textbf{M-BART} show lower similarity with other encoders, particularly at the lower and higher layers (\textbf{M-BERT}: $[1-3]$, $[10-12]$; \textbf{M-BART}: $[10, 11]$). Besides, \textbf{M-BART} and \textbf{M-BERT} demonstrate low pairwise similarity, being fairly different at the lower-to-higher layers $[3-11]$.

\section{Discussion}
\label{sec:discussion}

\paragraph{Three encoders exhibit similar behavior, but two other differ from them in capturing linguistic properties}
The most striking distinction based upon the results is that \textbf{M-BART} and \textbf{LABSE} exhibit different behavior for both languages, as opposed to \textbf{M-BERT}, \textbf{XLM-R}, and \textbf{MiniLM} (Section \ref{subsection:sup_probe}, \ref{subsec:neuron_analysis}). Specifically, \textbf{M-BART} generally tends to distribute the surface properties across all the layers, unlike other encoders where the information is specialized by the lower ones. The syntactic properties tend to be localized at the higher layers $[11-12]$ which is as well demonstrated in other Russian tasks (\textbf{Gapping}, \textbf{ImpersonalSent}). In contrast to other models that capture the semantic properties at the middle-to-higher layers, \textbf{LABSE} typically displays the knowledge at the lower-to-middle layers. The analysis of the understudied encoders contradicts the common findings on transformer models that syntactic information is stored at the middle layers, while semantic knowledge is most prominent at the higher layers \citep{rogers2020primer}. Therefore, there is still room for exploring transformer-based models that differ by architecture type, or by a set of pre-training objectives.

\paragraph{MiniLM inherits linguistic properties from M-BERT} Teacher models are usually compared with their distilled versions on a range of downstream tasks \citep{tsai2019small}, and little is explored on what language properties they bequeath to their students. We find that \textbf{MiniLM} is likely to mimic the behavior of \textbf{XLM-R} rather than of \textbf{M-BERT}, most probably due to using the same tokenization method. The similarity is most demonstrated by the individual neuron analysis (see Appendix \ref{appendix:neuron_analysis}) and \texttt{lincka} (see Figure \ref{fig:heatmap-lincka}). Along with that, the model receives comparative performance under layer-wise supervised probing (see Appendix \ref{probe_res}).

\paragraph{Surface and syntactic information is learned in a similar manner}
The probing curves demonstrate that the surface and semantic properties of the two languages are similarly distributed in the encoders. The surface properties are generally captured at the lower layers, and the pattern of the curves is decaying towards the output layer. The syntactic properties are predominantly inferred at the middle or higher layers, and the semantic tasks reveal a number of differences described in Section \ref{sec:results}. Notably, the results obtained under layer-wise supervised probing (Section \ref{subsection:sup_probe}) are supported by the individual neuron analysis (Section \ref{subsec:neuron_analysis}), and the representation-level analysis (Section \ref{subsec:corr_methods}).

\paragraph{Encoders may have similar distributions of neurons by task, but different individual neurons} 
Two neuron-level introspection methods allow drawing the following finding. Despite that top-neuron distributions in the encoder layers share similar patterns in the majority of probing tasks (Section \ref{subsec:neuron_analysis}), \texttt{maxcorr} induces high intra-encoder similarities and low inter-encoder similarities (Section \ref{subsec:corr_methods}). In other words, the neurons can similarly localize particular properties in the layers and yet behave differently across the models.

\section{Conclusion}
\label{sec:conclusion}
This paper introduces RuSentEval, an enhanced probing suite of 14 probing tasks that cover various linguistic phenomena of the Russian language. We explored five multilingual transformer encoders over the probing tasks on two typologically contrasting languages -- Russian and English. The experiments are conducted using a combination of complementary probing methods, including layer-wise supervised probing, individual neuron-analysis, neuron- and representation-level similarity measures. Particularly, the behavior of the encoders under probing classifiers is reflected in distributions of top neurons with respect to a task, and the similarity is supported by linear centered kernel alignment method. We found that despite the language distinctions, the surface and syntactic properties are learned in a fairly similar manner, and the semantic knowledge is captured differently in a majority of tasks. We believe that the findings make the ongoing studies on cross-lingual transfer even more promising, specifically from English to Russian or vice versa.
In contrast to prior works on how linguistic knowledge is represented in transformers, the analysis of the understudied models reveals that syntax and semantics can be differently represented across the layers.
Besides, we found that different encoders often have similar distributions of neurons that contribute most to a probing task, and yet differ under neuron-level similarities.
We also observed that distilled models inherit linguistic properties from their teachers, and receive comparative performance on a number of probing tasks.
An exciting direction for future work is to investigate the correlation between probing and high-level downstream tasks, in order to identify which linguistic properties anticipate the behavior of a model in action. 

\section*{Acknowledgments}
Ekaterina Artemova works within the framework of the HSE University Basic Research Program.

\bibliography{anthology,eacl2021}

\begin{thebibliography}{56}
\expandafter\ifx\csname natexlab\endcsname\relax\def\natexlab#1{#1}\fi

\bibitem[{Adi et~al.(2016)Adi, Kermany, Belinkov, Lavi, and
  Goldberg}]{adi2016fine}
Yossi Adi, Einat Kermany, Yonatan Belinkov, Ofer Lavi, and Yoav Goldberg. 2016.
\newblock {Fine-grained Analysis of Sentence Embeddings Using Auxiliary
  Prediction Tasks}.
\newblock \emph{arXiv preprint arXiv:1608.04207}.

\bibitem[{Alt et~al.(2020)Alt, Gabryszak, and Hennig}]{alt2020probing}
Christoph Alt, Aleksandra Gabryszak, and Leonhard Hennig. 2020.
\newblock {Probing Linguistic Features of Sentence-Level Representations in
  Relation Extraction}.
\newblock pages 1534--1545.

\bibitem[{Alzetta et~al.(2017)Alzetta, Dell’Orletta, Montemagni, and
  Venturi}]{alzetta2017dangerous}
Chiara Alzetta, Felice Dell’Orletta, Simonetta Montemagni, and Giulia
  Venturi. 2017.
\newblock {Dangerous Relations in Dependency Treebanks}.
\newblock pages 201--210.

\bibitem[{Anastasyev(2020)}]{anastasyev2020}
Daniil Anastasyev. 2020.
\newblock {Exploring Pretrained Models For Joint Morpho-syntactic Parsing of
  Russian}.

\bibitem[{Baerman(2007)}]{baerman2007syncretism}
Matthew Baerman. 2007.
\newblock {Syncretism}.
\newblock \emph{Language and Linguistics Compass}, 1(5):539--551.

\bibitem[{Belinkov and Glass(2019)}]{belinkov2019analysis}
Yonatan Belinkov and James Glass. 2019.
\newblock {Analysis Methods in Neural Language Processing: a Survey}.
\newblock \emph{Transactions of the Association for Computational Linguistics},
  7:49--72.

\bibitem[{Berg-Kirkpatrick et~al.(2012)Berg-Kirkpatrick, Burkett, and
  Klein}]{berg2012empirical}
Taylor Berg-Kirkpatrick, David Burkett, and Dan Klein. 2012.
\newblock {An Empirical Investigation of Statistical Significance in NLP}.
\newblock pages 995--1005.

\bibitem[{Bojanowski et~al.(2017)Bojanowski, Grave, Joulin, and
  Mikolov}]{bojanowski2017enriching}
Piotr Bojanowski, Edouard Grave, Armand Joulin, and Tomas Mikolov. 2017.
\newblock {Enriching Word Vectors with Subword Information}.
\newblock \emph{Transactions of the Association for Computational Linguistics},
  5:135--146.

\bibitem[{Burtsev et~al.(2018)Burtsev, Seliverstov, Airapetyan, Arkhipov,
  Baymurzina, Bushkov, Gureenkova, Khakhulin, Kuratov, Kuznetsov
  et~al.}]{burtsev2018deeppavlov}
Mikhail Burtsev, Alexander Seliverstov, Rafael Airapetyan, Mikhail Arkhipov,
  Dilyara Baymurzina, Nickolay Bushkov, Olga Gureenkova, Taras Khakhulin, Yurii
  Kuratov, Denis Kuznetsov, et~al. 2018.
\newblock {DeepPavlov: Open-source Library for Dialogue Systems}.
\newblock pages 122--127.

\bibitem[{Choenni and Shutova(2020{\natexlab{a}})}]{choenni2020cross}
Rochelle Choenni and Ekaterina Shutova. 2020{\natexlab{a}}.
\newblock {Cross-neutralising: Probing for Joint Encoding of Linguistic
  Information in Multilingual Models}.
\newblock \emph{arXiv e-prints}, pages arXiv--2010.

\bibitem[{Choenni and Shutova(2020{\natexlab{b}})}]{choenni2020does}
Rochelle Choenni and Ekaterina Shutova. 2020{\natexlab{b}}.
\newblock {What Does it Mean to be Language-agnostic? Probing Multilingual
  Sentence Encoders for Typological Properties}.
\newblock \emph{arXiv preprint arXiv:2009.12862}.

\bibitem[{Conneau et~al.(2020{\natexlab{a}})Conneau, Khandelwal, Goyal,
  Chaudhary, Wenzek, Guzm{\'a}n, Grave, Ott, Zettlemoyer, and
  Stoyanov}]{conneau2019unsupervised}
Alexis Conneau, Kartikay Khandelwal, Naman Goyal, Vishrav Chaudhary, Guillaume
  Wenzek, Francisco Guzm{\'a}n, {\'E}douard Grave, Myle Ott, Luke Zettlemoyer,
  and Veselin Stoyanov. 2020{\natexlab{a}}.
\newblock {Unsupervised Cross-lingual Representation Learning at Scale}.
\newblock pages 8440--8451.

\bibitem[{Conneau and Kiela(2018)}]{conneau2018senteval}
Alexis Conneau and Douwe Kiela. 2018.
\newblock {SentEval: An Evaluation Toolkit for Universal Sentence
  Representations}.

\bibitem[{Conneau et~al.(2018)Conneau, Kruszewski, Lample, Barrault, and
  Baroni}]{conneau2018you}
Alexis Conneau, Germ{\'a}n Kruszewski, Guillaume Lample, Lo{\"\i}c Barrault,
  and Marco Baroni. 2018.
\newblock {What you Can Cram into a Single Vector: Probing Sentence Embeddings
  for Linguistic Properties}.
\newblock \emph{arXiv preprint arXiv:1805.01070}.

\bibitem[{Conneau and Lample(2019)}]{conneau2019cross}
Alexis Conneau and Guillaume Lample. 2019.
\newblock {Cross-lingual Language Model Pre-training}.
\newblock pages 7059--7069.

\bibitem[{Conneau et~al.(2020{\natexlab{b}})Conneau, Rinott, Lample, Schwenk,
  Stoyanov, Williams, and Bowman}]{conneau2018xnli}
Alexis Conneau, Ruty Rinott, Guillaume Lample, Holger Schwenk, Ves Stoyanov,
  Adina Williams, and Samuel~R Bowman. 2020{\natexlab{b}}.
\newblock {XNLI: Evaluating Cross-lingual Sentence Representations}.
\newblock pages 2475--2485.

\bibitem[{Dalvi et~al.(2019)Dalvi, Durrani, Sajjad, Belinkov, Bau, and
  Glass}]{dalvi2019one}
Fahim Dalvi, Nadir Durrani, Hassan Sajjad, Yonatan Belinkov, Anthony Bau, and
  James Glass. 2019.
\newblock {What is One Grain of Sand in the Desert? Analyzing Individual
  Neurons in Deep NLP Models}.
\newblock 33:6309--6317.

\bibitem[{Devlin et~al.(2019)Devlin, Chang, Lee, and
  Toutanova}]{devlin2018bert}
Jacob Devlin, Ming-Wei Chang, Kenton Lee, and Kristina Toutanova. 2019.
\newblock {BERT: Pre-training of Deep Bidirectional Transformers for Language
  Understanding}.
\newblock pages 4171--4186.

\bibitem[{Droganova et~al.(2018)Droganova, Lyashevskaya, and
  Zeman}]{droganova2018data}
Kira Droganova, Olga Lyashevskaya, and Daniel Zeman. 2018.
\newblock {Data Conversion and Consistency of Monolingual Corpora: Russian UD
  Treebanks}.
\newblock pages 52--65.

\bibitem[{Durrani et~al.(2020)Durrani, Sajjad, Dalvi, and
  Belinkov}]{durrani2020analyzing}
Nadir Durrani, Hassan Sajjad, Fahim Dalvi, and Yonatan Belinkov. 2020.
\newblock {Analyzing Individual Neurons in Pre-trained Language Models}.
\newblock pages 4865--4880.

\bibitem[{Eger et~al.(2020)Eger, Daxenberger, and Gurevych}]{eger2020probe}
Steffen Eger, Johannes Daxenberger, and Iryna Gurevych. 2020.
\newblock {How to Probe Sentence Embeddings in Low-Resource Languages: On
  Structural Design Choices for Probing Task Evaluation}.
\newblock pages 108--118.

\bibitem[{Feng et~al.(2020)Feng, Yang, Cer, Arivazhagan, and
  Wang}]{feng2020language}
Fangxiaoyu Feng, Yinfei Yang, Daniel Cer, Naveen Arivazhagan, and Wei Wang.
  2020.
\newblock {Language-agnostic BERT Sentence Embedding}.
\newblock \emph{arXiv preprint arXiv:2007.01852}.

\bibitem[{Hu et~al.(2020)Hu, Ruder, Siddhant, Neubig, Firat, and
  Johnson}]{hu2020xtreme}
Junjie Hu, Sebastian Ruder, Aditya Siddhant, Graham Neubig, Orhan Firat, and
  Melvin Johnson. 2020.
\newblock {XTREME: A Massively Multilingual Multi-task Benchmark for Evaluating
  Cross-lingual Generalisation}.
\newblock pages 4411--4421.

\bibitem[{Kingma and Ba(2015)}]{kingma2014adam}
Diederik~P. Kingma and Jimmy Ba. 2015.
\newblock {Adam: A Method for Stochastic Optimization}.

\bibitem[{Kirov et~al.(2018)Kirov, Cotterell, Sylak-Glassman, Walther,
  Vylomova, Xia, Faruqui, Mielke, McCarthy, K{\"u}bler
  et~al.}]{kirov2018unimorph}
Christo Kirov, Ryan Cotterell, John Sylak-Glassman, G{\'e}raldine Walther,
  Ekaterina Vylomova, Patrick Xia, Manaal Faruqui, Sabrina~J Mielke, Arya~D
  McCarthy, Sandra K{\"u}bler, et~al. 2018.
\newblock {UniMorph 2.0: Universal Morphology}.

\bibitem[{Krasnowska-Kiera{\'s} and
  Wr{\'o}blewska(2019)}]{krasnowska2019empirical}
Katarzyna Krasnowska-Kiera{\'s} and Alina Wr{\'o}blewska. 2019.
\newblock {Empirical Linguistic Study of Sentence Embeddings}.
\newblock pages 5729--5739.

\bibitem[{Kudo and Richardson(2018)}]{kudo2018sentencepiece}
Taku Kudo and John Richardson. 2018.
\newblock {SentencePiece: A Simple and Language Independent Subword Tokenizer
  and Detokenizer for Neural Text Processing}.
\newblock pages 66--71.

\bibitem[{Liang et~al.(2020)Liang, Duan, Gong, Wu, Guo, Qi, Gong, Shou, Jiang,
  Cao et~al.}]{liang2020xglue}
Yaobo Liang, Nan Duan, Yeyun Gong, Ning Wu, Fenfei Guo, Weizhen Qi, Ming Gong,
  Linjun Shou, Daxin Jiang, Guihong Cao, et~al. 2020.
\newblock {XGLUE: A New Benchmark Datasetfor Cross-lingual Pre-training,
  Understanding and Generation}.
\newblock pages 6008--6018.

\bibitem[{Liu et~al.(2019)Liu, Gardner, Belinkov, Peters, and
  Smith}]{liu2019linguistic}
Nelson~F Liu, Matt Gardner, Yonatan Belinkov, Matthew~E Peters, and Noah~A
  Smith. 2019.
\newblock {Linguistic Knowledge and Transferability of Contextual
  Representations}.
\newblock pages 1073--1094.

\bibitem[{Liu et~al.(2020)Liu, Gu, Goyal, Li, Edunov, Ghazvininejad, Lewis, and
  Zettlemoyer}]{liu2020multilingual}
Yinhan Liu, Jiatao Gu, Naman Goyal, Xian Li, Sergey Edunov, Marjan
  Ghazvininejad, Mike Lewis, and Luke Zettlemoyer. 2020.
\newblock Multilingual denoising pre-training for neural machine translation.
\newblock \emph{Transactions of the Association for Computational Linguistics},
  8:726--742.

\bibitem[{Lyashevskaya and Sharov(2009)}]{lyashevskaya2009frequency}
Olga Lyashevskaya and Sergey Sharov. 2009.
\newblock {The Frequency Dictionary of Modern Russian Language}.
\newblock \emph{Azbukovnik, Moscow}.

\bibitem[{Mare{\v{c}}ek et~al.(2020)Mare{\v{c}}ek, Celikkanat, Silfverberg,
  Ravishankar, and Tiedemannb}]{marevcek2020multilingual}
David Mare{\v{c}}ek, Hande Celikkanat, Miikka Silfverberg, Vinit Ravishankar,
  and J{\"o}rg Tiedemannb. 2020.
\newblock {Are Multilingual Neural Machine Translation Models Better at
  Capturing Linguistic Features?}
\newblock \emph{The Prague Bulletin of Mathematical Linguistics},
  (115):143--162.

\bibitem[{de~Marneffe et~al.(2017)de~Marneffe, Grioni, Kanerva, and
  Ginter}]{de2017assessing}
Marie-Catherine de~Marneffe, Matias Grioni, Jenna Kanerva, and Filip Ginter.
  2017.
\newblock {Assessing the Annotation Consistency of the Universal Dependencies
  Corpora}.
\newblock pages 108--115.

\bibitem[{Michael et~al.(2020)Michael, Botha, and Tenney}]{michael2020asking}
Julian Michael, Jan~A Botha, and Ian Tenney. 2020.
\newblock {Asking without Telling: Exploring Latent Ontologies in Contextual
  Representations}.
\newblock \emph{arXiv preprint arXiv:2004.14513}.

\bibitem[{Nivre et~al.(2016)Nivre, De~Marneffe, Ginter, Goldberg, Hajic,
  Manning, McDonald, Petrov, Pyysalo, Silveira et~al.}]{nivre2016universal}
Joakim Nivre, Marie-Catherine De~Marneffe, Filip Ginter, Yoav Goldberg, Jan
  Hajic, Christopher~D Manning, Ryan McDonald, Slav Petrov, Sampo Pyysalo,
  Natalia Silveira, et~al. 2016.
\newblock {Universal Dependencies v1: A Multilingual Treebank Collection}.
\newblock pages 1659--1666.

\bibitem[{Pedregosa et~al.(2011)Pedregosa, Varoquaux, Gramfort, Michel,
  Thirion, Grisel, Blondel, Prettenhofer, Weiss, Dubourg
  et~al.}]{pedregosa2011scikit}
Fabian Pedregosa, Ga{\"e}l Varoquaux, Alexandre Gramfort, Vincent Michel,
  Bertrand Thirion, Olivier Grisel, Mathieu Blondel, Peter Prettenhofer, Ron
  Weiss, Vincent Dubourg, et~al. 2011.
\newblock {Scikit-learn: Machine learning in Python}.
\newblock \emph{the Journal of machine Learning research}, 12:2825--2830.

\bibitem[{Pires et~al.(2019)Pires, Schlinger, and
  Garrette}]{pires2019multilingual}
Telmo Pires, Eva Schlinger, and Dan Garrette. 2019.
\newblock {How Multilingual is Multilingual BERT?}
\newblock pages 4996--5001.

\bibitem[{Ponomareva et~al.(2019)Ponomareva, Droganova, Smurov, and
  Shavrina}]{ponomareva2019agrr}
Maria Ponomareva, Kira Droganova, Ivan Smurov, and Tatiana Shavrina. 2019.
\newblock {AGRR 2019: Corpus for Gapping Resolution in Russian}.
\newblock pages 35--43.

\bibitem[{Radford et~al.(2019)Radford, Wu, Child, Luan, Amodei, and
  Sutskever}]{radford2019language}
Alec Radford, Jeffrey Wu, Rewon Child, David Luan, Dario Amodei, and Ilya
  Sutskever. 2019.
\newblock {Language Models are Unsupervised Multitask Learners}.
\newblock \emph{OpenAI blog}, 1(8):9.

\bibitem[{Ravishankar et~al.(2019)Ravishankar, {\O}vrelid, and
  Velldal}]{ravishankar2019probing}
Vinit Ravishankar, Lilja {\O}vrelid, and Erik Velldal. 2019.
\newblock {Probing Multilingual Sentence Representations With XProbe}.
\newblock \emph{ACL 2019}, page 156.

\bibitem[{Rogers et~al.(2021)Rogers, Kovaleva, and
  Rumshisky}]{rogers2020primer}
Anna Rogers, Olga Kovaleva, and Anna Rumshisky. 2021.
\newblock {A Primer in BERTology: What we Know about how BERT Works}.
\newblock \emph{Transactions of the Association for Computational Linguistics},
  8:842--866.

\bibitem[{{\c{S}}ahin et~al.(2020){\c{S}}ahin, Vania, Kuznetsov, and
  Gurevych}]{csahin2020linspector}
G{\"o}zde~G{\"u}l {\c{S}}ahin, Clara Vania, Ilia Kuznetsov, and Iryna Gurevych.
  2020.
\newblock {Linspector: Multilingual Probing Tasks for Word Representations}.
\newblock \emph{Computational Linguistics}, 46(2):335--385.

\bibitem[{Sennrich et~al.(2016)Sennrich, Haddow, and
  Birch}]{sennrich2015neural}
Rico Sennrich, Barry Haddow, and Alexandra Birch. 2016.
\newblock {Neural Machine Translation of Rare Words with Subword Units}.
\newblock pages 1715--1725.

\bibitem[{Shavrina and Shapovalova(2017)}]{shavrina2017methodology}
Tatiana Shavrina and Olga Shapovalova. 2017.
\newblock {To the Methodology Corpus Construction for Machine Learning:
  ``TAIGA'' Syntax Tree Corpus and Parser}.
\newblock \emph{CORPUS LINGUISTICS, 2017}, page~78.

\bibitem[{Shi et~al.(2016)Shi, Padhi, and Knight}]{shi2016does}
Xing Shi, Inkit Padhi, and Kevin Knight. 2016.
\newblock {Does String-based Neural MT Learn Source Syntax?}
\newblock pages 1526--1534.

\bibitem[{Suau et~al.(2020)Suau, Zappella, and Apostoloff}]{suau2020finding}
Xavier Suau, Luca Zappella, and Nicholas Apostoloff. 2020.
\newblock {Finding Experts in Transformer Models}.
\newblock \emph{arXiv preprint arXiv:2005.07647}.

\bibitem[{Tenney et~al.(2019)Tenney, Xia, Chen, Wang, Poliak, McCoy, Kim,
  Van~Durme, Bowman, Das et~al.}]{tenney2019you}
Ian Tenney, Patrick Xia, Berlin Chen, Alex Wang, Adam Poliak, R~Thomas McCoy,
  Najoung Kim, Benjamin Van~Durme, Samuel~R Bowman, Dipanjan Das, et~al. 2019.
\newblock {What do you Learn from Context? Probing for Sentence Structure in
  Contextualized Word Representations}.
\newblock \emph{arXiv preprint arXiv:1905.06316}.

\bibitem[{Tsai et~al.(2019)Tsai, Riesa, Johnson, Arivazhagan, Li, and
  Archer}]{tsai2019small}
Henry Tsai, Jason Riesa, Melvin Johnson, Naveen Arivazhagan, Xin Li, and Amelia
  Archer. 2019.
\newblock {Small and Practical BERT Models for Sequence Labeling}.
\newblock pages 3623--3627.

\bibitem[{Vaswani et~al.(2017)Vaswani, Shazeer, Parmar, Uszkoreit, Jones,
  Gomez, Kaiser, and Polosukhin}]{vaswani2017attention}
Ashish Vaswani, Noam Shazeer, Niki Parmar, Jakob Uszkoreit, Llion Jones,
  Aidan~N Gomez, {\L}ukasz Kaiser, and Illia Polosukhin. 2017.
\newblock {Attention is All You Need}.
\newblock pages 5998--6008.

\bibitem[{Voita and Titov(2020)}]{voita2020information}
Elena Voita and Ivan Titov. 2020.
\newblock {Information-Theoretic Probing with Minimum Description Length}.
\newblock \emph{arXiv preprint arXiv:2003.12298}.

\bibitem[{Wang et~al.(2019)Wang, Pruksachatkun, Nangia, Singh, Michael, Hill,
  Levy, and Bowman}]{wang2019superglue}
Alex Wang, Yada Pruksachatkun, Nikita Nangia, Amanpreet Singh, Julian Michael,
  Felix Hill, Omer Levy, and Samuel Bowman. 2019.
\newblock {SuperGLUE: A Stickier Benchmark for General-purpose Language
  Understanding Systems}.
\newblock pages 3266--3280.

\bibitem[{Wang et~al.(2020)Wang, Wei, Dong, Bao, Yang, and
  Zhou}]{wang2020minilm}
Wenhui Wang, Furu Wei, Li~Dong, Hangbo Bao, Nan Yang, and Ming Zhou. 2020.
\newblock {Minilm: Deep Self-attention Distillation for Task-agnostic
  Compression of Pre-trained Transformers}.
\newblock \emph{arXiv preprint arXiv:2002.10957}.

\bibitem[{Wenzek et~al.(2020)Wenzek, Lachaux, Conneau, Chaudhary, Guzm{\'a}n,
  Joulin, and Grave}]{wenzek2019ccnet}
Guillaume Wenzek, Marie-Anne Lachaux, Alexis Conneau, Vishrav Chaudhary,
  Francisco Guzm{\'a}n, Armand Joulin, and {\'E}douard Grave. 2020.
\newblock {CCNet: Extracting High Quality Monolingual Datasets from Web Crawl
  Data}.
\newblock pages 4003--4012.

\bibitem[{Wolf et~al.(2019)Wolf, Debut, Sanh, Chaumond, Delangue, Moi, Cistac,
  Rault, Louf, Funtowicz et~al.}]{wolf2019huggingface}
Thomas Wolf, Lysandre Debut, Victor Sanh, Julien Chaumond, Clement Delangue,
  Anthony Moi, Pierric Cistac, Tim Rault, R{\'e}mi Louf, Morgan Funtowicz,
  et~al. 2019.
\newblock {HuggingFace's Transformers: State-of-the-art Natural Language
  Processing}.
\newblock \emph{ArXiv}, pages arXiv--1910.

\bibitem[{Wu et~al.(2020)Wu, Belinkov, Sajjad, Durrani, Dalvi, and
  Glass}]{wu2020similarity}
John Wu, Yonatan Belinkov, Hassan Sajjad, Nadir Durrani, Fahim Dalvi, and James
  Glass. 2020.
\newblock {Similarity Analysis of Contextual Word Representation Models}.
\newblock pages 4638--4655.

\bibitem[{Zou and Hastie(2005)}]{zou2005regularization}
Hui Zou and Trevor Hastie. 2005.
\newblock {Regularization and Variable Selection via the Elastic Net}.
\newblock \emph{Journal of the royal statistical society: series B (statistical
  methodology)}, 67(2):301--320.

\end{thebibliography}
\bibliographystyle{acl_natbib}

\appendix
\newpage
\onecolumn
\section{Examples from RuSentEval Tasks}
\label{appendix:examples}
Table~\ref{tab:appendix_data_description} provides with examples from RuSentEval tasks.

\begin{table*}[htp!]
\centering

\begin{tabular}{c|l|c}
\toprule
\bf Task & \bf Example & \bf Label \\
\midrule

\multirow{2}{*}{{\bf ConjType}} & On otmetil , \textbf{chto} podobnyye progulki nebezopasny . &  \multirow{2}{*}{\textbf{SCONJ}} \\
& \textit{'He noted \textbf{that} such walks are unsafe.'} &  \\
\midrule

\multirow{2}{*}{{\bf Gapping}} & Ya yezdila dvazhdy , sestra – trizhdy . & \multirow{2}{*}{\textbf{1}} \\
& \textit{'I went [there] twice, my sister [went there] three times'}&  \\ 
\midrule

\multirow{2}{*}{{\bf ImpersonalSent}} & Rabotal takzhe kak kontsertmeyster i lektor . &  \multirow{2}{*}{\bf{0}} \\
& \textit{'[He] also worked as an accompanist and lecturer.'} &  \\ 
\midrule

\multirow{2}{*}{{\bf NShift}} & Kogda etogo \textbf{poluchilos' ne} , on ubezhal .&  \multirow{2}{*}{\bf{I}}\\
& \textit{'When it \textbf{work out didn't}, he ran away'}&  \\ 
\midrule

\multirow{2}{*}{{\bf ObjGender}} & Rossiyskiy duet dopustil odnu \textbf{oshibku} . & \multirow{2}{*}{\bf{F}} \\
&\textit{'The Russian duo made one \textbf{mistake}.'} &  \\
\midrule

\multirow{2}{*}{{\bf ObjNumber}} & Serial poluchil neskol'ko prestizhnykh \textbf{nagrad} . & \multirow{2}{*}{\bf{NNS}} \\
&\textit{'The series has received several prestigious \textbf{awards}.'} &  \\
\midrule

\multirow{2}{*}{{\bf PA}} & On nikak ne \textbf{ob"yasnil} svoyu pozitsiyu . & \multirow{2}{*}{\bf{PERF}} \\
&\textit{'He did not \textbf{explain} his position in any way .'} &  \\
\midrule

\multirow{3}{*}{{\bf PT}} & Molodyye spetsialisty \textbf{poluchayut}  yezhemesyachnuyu doplatu k &  \multirow{3}{*}{\bf{PRES}} \\
& zarplate . & \\
&\textit{'Young professionals \textbf{receive} a monthly supplement to their salary .'} &  \\
\midrule

\multirow{3}{*}{{\bf PV} } & Srok vozmozhnoy prem'yery lenty poka ne \textbf{nazyvayetsya} &   \multirow{3}{*}{\bf{PASS}}\\
&\textit{'The date of a possible premiere of the film has not yet been} & \\
& \textit{\textbf{announced}.'} & \\
\midrule

\multirow{2}{*}{{\bf SentLen}} & Ya ne videla boleye zlogo cheloveka . & \multirow{2}{*}{\bf{0}} \\
 &\textit{'I haven't seen a more angry man.'} &  \\
\midrule

\multirow{2}{*}{{\bf SubjGender}} & \textbf{On} nosit beluyu dlinnuyu rubashku i dlinnyye seryye bryuki. & \multirow{2}{*}{\bf{M}} \\
&\textit{'\textbf{He} wears a white long shirt and long grey trousers.'} &  \\
\midrule

\multirow{3}{*}{{\bf SubjNumber}} & \textbf{On} byl lyubimtsem vsey moskovskoy i peterburgskoy aristokratii . & \multirow{3}{*}{\bf{NN}}  \\
&\textit{'\textbf{He} was a favorite of the entire Moscow and St. Petersburg} &  \\
& \textit{aristocracy.'} & \\
\midrule

\multirow{2}{*}{{\bf TreeDepth}} & I vot v pervuyu ochered' my khoteli by pogovorit' ob etom . & \multirow{2}{*}{\bf{5}} \\
& \textit{'And first of all, we would like to talk about this.'}&  \\ 
\midrule

\multirow{2}{*}{{\bf WC}} & Proshluyu noch' ya sovsem ne \textbf{spal} . &  \bf{spat'} \\ 
& \textit{'I didn't \textbf{sleep} at all last night.'} & \textit{'to sleep'}  \\ 
\bottomrule

\end{tabular}
\caption{Examples from RuSentEval tasks.}
\label{tab:appendix_data_description}
\end{table*}

\newpage 

\section{Layer-wise Supervised Probing}
\label{probe_res}
\subsection{Results on All Tasks}

The results reported in the main body of the paper are obtained with Logistic Regression classifier over the shared tasks (see Section \ref{subsection:sup_probe}). We present detailed results for both linear and non-linear classifiers on all Russian and English tasks in Tables \ref{tab:ru_logreg}--\ref{tab:en_mlp}. Tables \ref{tab:baseline_ru_logreg}--\ref{tab:baseline_ru_mlp_1} show the results of the baselines. Table \ref{tab:data_description} outlines statistical description of the tasks.

\begin{table*}[h]
\centering
\begin{tabular}{c|c|c|c|c|c}
\toprule
\textbf{Probing Task} &  \textbf{M-BERT} &  \textbf{LABSE} &  \textbf{XLM-R} &  \textbf{MiniLM} &  \textbf{M-BART} \\
\midrule
\textbf{ConjType}       &   98.8 [7] &  \colorbox{lightgray}{\textbf{99.3 [4]}} &   \colorbox{lightgray}{\textbf{99.3 [6]}} &   98.6 [5] &   98.8 [7] \\
\textbf{Gapping}        &   85.2 [7] &  89.7 [8] &   \colorbox{lightgray}{\textbf{94.1 [8]}} &   91.1 [9] &  85.6 [12] \\
\textbf{ImpersonalSent} &   91.6 [7] &  92 [6] &   \colorbox{lightgray}{\textbf{92.6 [4]}} &   88.4 [6] &  85.7 [12] \\
\textbf{NShift}         &  81.8 [10] &  82.6 [5] &   \colorbox{lightgray}{\textbf{86.9 [9]}} &   80.5 [9] &  78.6 [12] \\
\textbf{ObjGender}      &   70.1 [6] &  70.4 [2] &   69.4 [5] &   64.1 [9] &   \colorbox{lightgray}{\textbf{71.8 [1]}} \\
\textbf{ObjNumber}      &   82.8 [6] &  82.5 [2] &  \colorbox{lightgray}{\textbf{83.7 [10]}} &  77.8 [10] &   81.5 [7] \\
\textbf{PA}             &   91.2 [6] &  93.8 [4] &   94.4 [5] &   89.4 [5] &  \colorbox{lightgray}{\textbf{95.9 [10]}} \\
\textbf{PT}             &   99.5 [8] &  \colorbox{lightgray}{\textbf{99.8 [5]}} &   \colorbox{lightgray}{\textbf{99.8 [5]}} &   98.2 [7] &   99.6 [7] \\
\textbf{PV}             &   77.5 [5] &  76.3 [5] &   76.8 [5] &   71.4 [3] &   \colorbox{lightgray}{\textbf{77.7 [3]}} \\
\textbf{SentLen}        &   91.3 [2] &  93.3 [1] &   94.5 [2] &   94.1 [2] &   \colorbox{lightgray}{\textbf{96.2 [4]}} \\
\textbf{SubjGender}     &   79.1 [9] &  79.2 [2] &  \colorbox{lightgray}{\textbf{79.4 [11]}} &  78.7 [10] &   77.7 [7] \\
\textbf{SubjNumber}    &   90.5 [7] &  92.9 [3] &  \colorbox{lightgray}{\textbf{94.9 [11]}} &  94.2 [12] &  93.1 [10] \\
\textbf{TreeDepth}      &   44.7 [6] &  46.1 [4] &   \colorbox{lightgray}{\textbf{46.5 [5]}} &   44.8 [7] &  45.8 [11] \\
\textbf{WC}             &   84.8 [2] &  85.8 [1] &   82.6 [1] &   72.8 [1] &   \colorbox{lightgray}{\textbf{88.0 [1]}} \\
\bottomrule
\end{tabular}
\caption{Results of Logistic Regression classifier by the encoder for RuSentEval tasks.}
\label{tab:ru_logreg}

\vspace*{1 cm}

\centering
\begin{tabular}{c|c|c|c|c|c}
\toprule
\textbf{Probing Task} &  \textbf{M-BERT} &  \textbf{LABSE} &  \textbf{XLM-R} &  \textbf{MiniLM} &  \textbf{M-BART} \\
\midrule
\textbf{BShift}     &  84.8 [8] &   84.4 [5] &  \colorbox{lightgray}{\textbf{85.7 [10]}} &  79.3 [8] &  83.8 [12] \\
\textbf{CoordInv}   &  66.0 [8] &   68.9 [8] &   68.6 [8] &  63.3 [8] &  \colorbox{lightgray}{\textbf{69.4 [12]}} \\
\textbf{ObjNumber}  & \colorbox{lightgray}{\textbf{86.2 [6]}} &   85.4 [3] &   86.0 [8] &  85.2 [6] &   85.9 [9] \\
\textbf{SOMO}       &  57.4 [8] &   60.8 [7] &   60.0 [8] &  56.1 [9] &  \colorbox{lightgray}{\textbf{62.3 [12]}} \\
\textbf{SentLen}    &  96.3 [2] &   96.6 [1] &   95.8 [2] &  96.1 [3] &   \colorbox{lightgray}{\textbf{97.3 [3]}} \\
\textbf{SubjNumber} &  87.8 [7] &  \colorbox{lightgray}{\textbf{90.7 [12]}} &  86.9 [10] &  85.6 [6] &   87.3 [9] \\
\textbf{Tense}      &  88.9 [8] &   88.8 [6] &   88.8 [9] &  87.3 [5] &   \colorbox{lightgray}{\textbf{89.1 [9]}} \\
\textbf{TopConst}   &  \colorbox{lightgray}{\textbf{88 [6]}} &   79.9 [5] &   78.5 [5] &  76.5 [5] &   79.5 [8] \\
\textbf{TreeDepth}  &  41.2 [5] &  \colorbox{lightgray}{\textbf{42.7 [5]}} &   41.8 [7] &  40.9 [7] &  41.2 [12] \\
\textbf{WC}         &  92.6 [1] &   93.7 [1] &   89.8 [1] &  82.3 [1] &  \colorbox{lightgray}{\textbf{93.8 [1]}} \\
\bottomrule
\end{tabular}
\caption{Results of Logistic Regression classifier by the encoder for SentEval tasks.}
\label{tab:en_logreg}
\end{table*}

\begin{table*}
\centering
\begin{tabular}{c|c|c|c|c|c}
\toprule
\textbf{Probing Task} &  \textbf{M-BERT} &  \textbf{LABSE} &  \textbf{XLM-R} &  \textbf{MiniLM} &  \textbf{M-BART} \\
\midrule
\textbf{ConjType}       &   98.6 [5] &   \colorbox{lightgray}{\textbf{99.4 [4]}} &   99.2 [5] &   98.9 [5] &   98.8 [7] \\
\textbf{Gapping}        &  89.7 [10] &   90.0 [9] &   \colorbox{lightgray}{\textbf{96.0 [8]}} &  92.0 [11] &   83.1 [9] \\
\textbf{ImpersonalSent} &    \colorbox{lightgray}{\textbf{93.6 [7]}} &   90.9 [5] &   92.4 [7] &   88.7 [7] &   89.4 [9] \\
\textbf{NShift}         &   81.5 [9] &   82.7 [5] &    \colorbox{lightgray}{\textbf{87.6 [9]}} &  81.1 [10] &  78.8 [12] \\
\textbf{ObjGender}      &   69.1 [6] &   70.1 [2] &   69.5 [9] &  65.1 [10] &    \colorbox{lightgray}{\textbf{72.2 [1]}} \\
\textbf{ObjNumber}      &   83.9 [6] &   82.5 [2] &   \colorbox{lightgray}{\textbf{84.8 [10]}} &  78.7 [10] &   83.0 [7] \\
\textbf{PA}             &   90.9 [7] &   93.5 [5] &   94.6 [5] &   89.7 [5] &    \colorbox{lightgray}{\textbf{95.5 [8]}} \\
\textbf{PT}             &   99.4 [4] &   \colorbox{lightgray}{\textbf{99.9 [10]}} &   99.8 [5] &   98.4 [6] &   99.6 [9] \\
\textbf{PV}             &   77.7 [4] &   76.5 [5] &   78.4 [4] &   72.5 [4] &    \colorbox{lightgray}{\textbf{82.2 [1]}} \\
\textbf{SentLen}        &   93.5 [2] &   95.2 [1] &   97.1 [2] &   96.7 [1] &    \colorbox{lightgray}{\textbf{98.2 [5]}} \\
\textbf{SubjGender}     &   79.5 [9] &   80.0 [2] &   \colorbox{lightgray}{\textbf{81.0 [11]}} &  80.2 [12] &   78.1 [7] \\
\textbf{SubjNumber}    &   90.3 [5] &   93.0 [3] &   \colorbox{lightgray}{\textbf{96.3 [12]}} &  95.8 [11] &   94.5 [7] \\
\textbf{TreeDepth}      &   43.6 [6] &   45.4 [4] &   44.8 [7] &    \colorbox{lightgray}{\textbf{46.7 [8]}} &   46.0 [7] \\
\textbf{WC}             &   80.8 [3] &   82.7 [1] &   78.5 [1] &   69.9 [1] &    \colorbox{lightgray}{\textbf{84.4 [1]}} \\
\bottomrule
\end{tabular}
\caption{Results of MLP classifier by the encoder for RuSentEval tasks.}
\label{tab:ru_mlp}
\vspace*{1 cm}
\centering
\begin{tabular}{c|c|c|c|c|c}
\toprule
\textbf{Probing Task} &  \textbf{M-BERT} &  \textbf{LABSE} &  \textbf{XLM-R} &  \textbf{MiniLM} &  \textbf{M-BART} \\
\midrule
\textbf{BShift}     &   83.1 [8] &   84.7 [6] &   \colorbox{lightgray}{\textbf{85.8 [9]}} &  79.4 [7] &  84.4 [12] \\
\textbf{CoordInv}   &   65.2 [8] &   68.1 [8] &   \colorbox{lightgray}{\textbf{68.8 [8]}} &  63.7 [8] &  67.8 [10] \\
\textbf{ObjNumber}  &   86.5 [6] &  86.4 [10] &   86.4 [8] &  85.0 [6] &   \colorbox{lightgray}{\textbf{86.8 [8]}} \\
\textbf{SOMO}       &   56.5 [8] &   60.5 [7] &   58.8 [8] &  54.8 [9] &  \colorbox{lightgray}{\textbf{61.7 [11]}} \\
\textbf{SentLen}    &   97.0 [2] &   98.4 [1] &   98.0 [3] &  98.5 [1] &   \colorbox{lightgray}{\textbf{98.8 [4]}} \\
\textbf{SubjNumber} &  86.5 [10] &  \colorbox{lightgray}{\textbf{90.9 [11]}} &   86.6 [6] &  86.0 [6] &   87.5 [9] \\
\textbf{Tense}      &   89.2 [9] &   89.0 [6] &  88.5 [10] &  87.9 [5] &   \colorbox{lightgray}{\textbf{89.4 [9]}} \\
\textbf{TopConst}   &   \colorbox{lightgray}{\textbf{82.0 [7]}} &   80.6 [5] &   79.5 [5] &  77.8 [6] &   80.6 [8] \\
\textbf{TreeDepth}  &   41.9 [6] &   43.1 [5] &   43.2 [7] &  42.3 [6] & \colorbox{lightgray}{\textbf{45.3 [10]}} \\
\textbf{WC}         &   91.2 [1] &   92.7 [1] &   88.9 [1] &  80.3 [1] &   \colorbox{lightgray}{\textbf{93.0 [1]}} \\

\bottomrule
\end{tabular}
\caption{Results of MLP classifier by encoder for each SentEval task.}
\label{tab:en_mlp}
\end{table*}

\begin{table*}
\centering
\begin{tabular}{c|c|c|c|c}
\toprule
\textbf{Probing Task} &  \textbf{fastText} &  \textbf{TF-IDF Char} &  \textbf{TF-IDF BPE} &  \textbf{TF-IDF SP} \\
\midrule
\textbf{ConjType}       &  88.1 & \colorbox{lightgray}{\textbf{96.9}} & 95.4 & 95.5 \\
\textbf{Gapping}        &  \colorbox{lightgray}{\textbf{84.1}} & 82.7 & 80.4 & 80.6 \\
\textbf{ImpersonalSent} & \colorbox{lightgray}{\textbf{78.7}} & 69.4 & 53.8 &  56.3 \\
\textbf{NShift}         & \colorbox{lightgray}{\textbf{53.2}} & 53.0 & 51.0 &  50.5 \\
\textbf{ObjGender}      & 70.1 & \colorbox{lightgray}{\textbf{71.0}} & 35.4 & 38.9 \\
\textbf{ObjNumber}      & \colorbox{lightgray}{\textbf{82.3}} &  76.4 & 56.8 & 55.0 \\
\textbf{PA}             & \colorbox{lightgray}{\textbf{90.8}} &  80.7 & 53.4 & 54.2 \\
\textbf{PT}             & 95.1 & \colorbox{lightgray}{\textbf{97.7}} & 53.8 &  53.7 \\
\textbf{PV}             & 69.2 & \colorbox{lightgray}{\textbf{78.2}} &  36.0 &   37.0 \\
\textbf{SentLen}        & 40.4 &  \colorbox{lightgray}{\textbf{64.0}} &  42.9 &  42.2 \\
\textbf{SubjGender}     &  \colorbox{lightgray}{\textbf{78.7}} &  74.4 &   34.8 &  38.0 \\
\textbf{SubjNumber}    &  \colorbox{lightgray}{\textbf{95.0}} &  90.4 &  63.7 & 64.4 \\
\textbf{TreeDepth}      & \colorbox{lightgray}{\textbf{35.7}} & 32.7 & 26.5 & 24.8 \\
\textbf{WC}             &  \colorbox{lightgray}{\textbf{70.8}}&  49.2 &  22.0 &   13.0 \\

\bottomrule
\end{tabular}
\caption{Results of Logistic Regression classifier by the baseline feature for each RuSentEval task.}
\label{tab:baseline_ru_logreg}
\vspace*{1 cm}
\centering
\begin{tabular}{c|c|c|c|c}
\toprule
\textbf{Probing Task} &  \textbf{fastText} &  \textbf{TF-IDF Char} &  \textbf{TF-IDF BPE} &  \textbf{TF-IDF SP} \\
\midrule
\textbf{BShift}         &  50.0 &  \colorbox{lightgray}{\textbf{51.1}} &   49.9 &  50.1 \\
\textbf{CoordInv}      &  52.2 &  \colorbox{lightgray}{\textbf{54.9}} &   50.2 &  50.1 \\
\textbf{ObjNumber}      & 72.8 &  \colorbox{lightgray}{\textbf{79.4}} &   68.1 &  69.0 \\
\textbf{SOMO}      &  49.9 & 49.9 &  \colorbox{lightgray}{\textbf{50.4}} &  49.7 \\
\textbf{SentLen}        &  \colorbox{lightgray}{\textbf{65.2}} &  54.1 &  42.3 &  44.6 \\
\textbf{SubjNumber}    & 76.6 &  \colorbox{lightgray}{\textbf{79.2}} &  68.1 & 71.6 \\
\textbf{Tense}      &  81.2 &   \colorbox{lightgray}{\textbf{84.2}} &   70.8 &  74.2 \\
\textbf{TopConst}      &   \colorbox{lightgray}{\textbf{59.8}} & 58.3 &  23.0 &  23.4 \\
\textbf{TreeDepth}      &   \colorbox{lightgray}{\textbf{30.0}} & 28.3 &  23.3 & 23.2 \\
\textbf{WC}             & 18.1 &  \colorbox{lightgray}{\textbf{47.3}} & 20.0 & 24.0 \\

\bottomrule
\end{tabular}
\caption{Results of Logistic Regression classifier by the baseline feature for each SentEval task.}
\label{tab:baseline_en_logreg}
\end{table*}

\begin{table*}
\centering
\begin{tabular}{c|c|c|c|c}
\toprule
\textbf{Probing Task} &  \textbf{fastText} &  \textbf{TF-IDF Char} &  \textbf{TF-IDF BPE} &  \textbf{TF-IDF SP} \\
\midrule

\textbf{ConjType}       &  88.4 &  \colorbox{lightgray}{\textbf{97.3}} &   95.6 &  95.5 \\
\textbf{Gapping}        &  82.7 &  \colorbox{lightgray}{\textbf{86.1}} &  80.2 &  68.8 \\
\textbf{ImpersonalSent} & \colorbox{lightgray}{\textbf{78.6}} &  70.5 &   52.9 &  56.6 \\
\textbf{NShift}         &  \colorbox{lightgray}{\textbf{52.7}} &  \colorbox{lightgray}{\textbf{52.7}} &   50.0 &  50.6 \\
\textbf{ObjGender}      &  70.0 & \colorbox{lightgray}{\textbf{70.9}} &   35.1 & 37.2 \\
\textbf{ObjNumber}      &  \colorbox{lightgray}{\textbf{82.8}} &  77.2 &   56.6 &  54.8 \\
\textbf{PA}             &  \colorbox{lightgray}{\textbf{91.2}} &  80.9 &  51.8 &   53.5 \\
\textbf{PT}             &  96.0 &  \colorbox{lightgray}{\textbf{97.6}} &   54.4 &  54.1 \\
\textbf{PV}             &  68.5 &  \colorbox{lightgray}{\textbf{78.5}} &  35.3 &   36.8 \\
\textbf{SentLen}        &  42.4 &  \colorbox{lightgray}{\textbf{73.7}} &  42.7 &  42.4 \\
\textbf{SubjGender}     &  \colorbox{lightgray}{\textbf{80.0}} &  75.2 &   34.0 &  38.8 \\
\textbf{SubjNumber}    &  \colorbox{lightgray}{\textbf{96.2}} &  90.8 &  61.8 & 64.4 \\
\textbf{TreeDepth}      &  29.5 &  \colorbox{lightgray}{\textbf{35.6}} &  32.8 &   23.9 \\
\textbf{WC}             &  \colorbox{lightgray}{\textbf{71.2}} &  53.8 &  20.0 &   11.0 \\

\bottomrule
\end{tabular}
\caption{Results of MLP classifier by the baseline feature for each RuSentEval task.}
\label{tab:baseline_ru_mlp_1}

\vspace*{1 cm}

\centering
\begin{tabular}{c|c|c|c|c}
\toprule
\textbf{Probing Task} &  \textbf{fastText} &  \textbf{TF-IDF Char} &  \textbf{TF-IDF BPE} &  \textbf{TF-IDF SP} \\
\midrule
\textbf{BShift}         &  48.2 & \colorbox{lightgray}{\textbf{50.6}} & 50.0 &  49.3 \\
\textbf{CoordInv}      &  50.1 & \colorbox{lightgray}{\textbf{54.0}} &   50.0 &  51.7 \\
\textbf{ObjNumber}      & 70.9 &  \colorbox{lightgray}{\textbf{77.1}} &   68.1 &  70.0 \\
\textbf{SOMO}      &  50.1 & 49.9 &  \colorbox{lightgray}{\textbf{50.2}} &  \colorbox{lightgray}{\textbf{50.2}} \\
\textbf{SentLen}        &  49.1 &  \colorbox{lightgray}{\textbf{62.5}} &  41.8 &  43.5 \\
\textbf{SubjNumber}    & 72.8 &  \colorbox{lightgray}{\textbf{80.5}} &  66.4 & 71.3 \\
\textbf{Tense}      & 74.7 &   \colorbox{lightgray}{\textbf{85.0}} &   70.5 &  73.8 \\
\textbf{TopConst}      &   58.0 & \colorbox{lightgray}{\textbf{59.7}} &  22.2 &  23.0 \\
\textbf{TreeDepth}      & 23.0  & \colorbox{lightgray}{\textbf{29.5}} &  23.0 & 22.1 \\
\textbf{WC}             & \colorbox{lightgray}{\textbf{63.3}} &  54.4 & 18.0 & 22.0 \\

\bottomrule
\end{tabular}
\caption{Results of MLP classifier by the baseline feature for each SentEval task.}
\label{tab:baseline_en_mlp}
\end{table*}

\begin{table*}
\centering
\begin{tabular}{|c|cccc|cccc|}
\toprule
\multirow{2}{*}{} & \multicolumn{4}{c|}{\bf RuSentEval} & \multicolumn{4}{c|}{\bf SentEval} \\
& Train & Dev & Test & Overall & Train & Dev & Test & Overall \\
\specialrule{.1em}{.05em}{.05em} 
 & \multicolumn{4}{c|}{\textbf{SentLen}} & \multicolumn{4}{c|}{\textbf{SentLen}} \\
\midrule

sample size & 100k & 10k & 10k & 120k & 100k & 10k & 10k & 120k \\
tokens & 1.45kk & 144.8k & 144.8k & 1.74kk & 1.66kk & 165.4k & 165.4k & 1.99kk \\
unique tokens & 116.7k & 34.0k & 33.7k & 126.5k & 34.8k & 9.6k & 10.0k & 36.8k \\
tokens/sentence & 14.47 & 14.48 & 14.48 & 14.47 & 16.59 & 16.54 & 16.55 & 16.59 \\
label distribution & \multicolumn{4}{c|}{16.7/16.7/16.7/16.7/16.7/16.7} & \multicolumn{4}{c|}{16.7/16.7/16.7/16.7/16.7/16.7} \\
\midrule
& \multicolumn{4}{c|}{\textbf{WC}} & \multicolumn{4}{c|}{\textbf{WC}} \\
\midrule
sample size & 100k & 10k & 10k & 120k & 100k & 10k & 10k & 120k \\
tokens & 1.19kk & 117.9k & 118.9k & 1.43kk & 1.5kk & 149.2k & 149.8k & 1.8kk \\
unique tokens & 106.6k & 30.6k & 30.8k & 115.7k & 37.4k & 13.4k & 13.4k & 40.1k \\
tokens/sentence & 11.89 & 11.79 & 11.89 & 11.88 & 15.02 & 14.92 & 14.98 & 15.00 \\
label distribution & \multicolumn{4}{c|}{0.1/label, 1000 labels} & \multicolumn{4}{c|}{0.1/label, 1000 labels} \\
\hline

& \multicolumn{4}{c|}{\textbf{NShift}} & \multicolumn{4}{c|}{\textbf{BShift}} \\
\midrule
sample size & 100k & 10k & 10k & 120k & 100k & 10k & 10k & 120k \\
tokens & 1.49kk & 125.0k & 127.4k & 1.74kk & 1.38kk & 137.4k  & 136.4k & 1.65kk \\
unique tokens & 138.2k & 29.2k & 29.6k & 146.9k & 36.2k & 12.8k & 12.7k & 38.7k \\
tokens/sentence & 14.88 & 12.74 & 12.51 & 14.51 & 13.78 & 13.74 & 13.64 & 13.77 \\
label distribution & \multicolumn{4}{c|}{50/50} & \multicolumn{4}{c|}{50/50} \\
\midrule

& \multicolumn{4}{c|}{\textbf{SubjNumber}} & \multicolumn{4}{c|}{\textbf{SubjNumber}} \\
\midrule
sample size & 100k & 10k & 10k & 120k & 100k & 10k & 10k & 120k \\
tokens & 1.04kk & 99.9k & 103.2k & 1.25kk & 1.41kk & 140.3k & 141.7k & 1.7kk \\
unique tokens & 100.3k & 21.7k & 23.1k & 108.3k & 38.5k & 14.4k & 14.5k & 41.3k \\
tokens/sentence & 10.42 & 9.99 & 10.32 & 10.38 & 14.14 & 14.03 & 14.17 & 14.13 \\
label distribution & \multicolumn{4}{c|}{50/50} & \multicolumn{4}{c|}{50/50} \\
\midrule

& \multicolumn{4}{c|}{\textbf{ObjNumber}} & \multicolumn{4}{c|}{\textbf{ObjNumber}} \\
\midrule
sample size & 100k & 10k & 10k & 120k & 100k & 10k & 10k & 120k \\
tokens & 946.7k & 103.3k & 100.2k & 1.15kk & 1.4kk & 140.5k & 139.9k & 1.68kk \\
unique tokens & 86.0k & 23.8k & 22.8k & 95.7k & 38.0k & 14.3k & 13.9k & 40.9k \\
tokens/sentence & 9.47 & 10.33 & 10.02 & 9.59 & 13.96 & 14.05 & 13.99 & 13.97 \\
label distribution & \multicolumn{4}{c|}{50/50} & \multicolumn{4}{c|}{50/50} \\
\midrule
& \multicolumn{4}{c|}{\textbf{PT}} & \multicolumn{4}{c|}{\textbf{Tense}} \\
\hline
sample size & 100k & 10k & 10k & 120k & 100k & 10k & 10k & 120k \\
tokens & 1.13kk & 112.8k & 113.0k & 1.36kk & 1.32kk & 131.1k & 129.6k & 1.58kk \\
unique tokens & 114.5k & 26.8k & 26.7k & 126.6k & 35.9k & 13.1k & 13.2k & 38.6k \\
tokens/sentence & 11.30 & 11.28 & 11.30 & 11.30 & 13.20 & 13.11 & 12.96 & 13.17 \\
label distribution & \multicolumn{4}{c|}{50/50} & \multicolumn{4}{c|}{50/50} \\
\midrule

& \multicolumn{4}{c|}{\textbf{TreeDepth}} & \multicolumn{4}{c|}{\textbf{TreeDepth}} \\
\midrule
sample size & 100k & 10k & 10k & 120k & 100k & 10k & 10k & 120k \\
tokens & 1.57kk & 157.2k & 157.6k & 1.88kk & 1.35kk & 135.0k & 134.7k & 1.62kk \\
unique tokens & 150.3k & 38.4k & 38.3k & 165.7k & 34.8k & 12.5k & 12.6k & 37.1k \\
tokens/sentence & 15.72 & 15.72 & 15.76 & 15.73 & 13.47 & 13.50 & 13.47 & 13.47 \\
label distribution & \multicolumn{4}{c|}{13.6/22.6/32.9/20.7/10.3} & \multicolumn{4}{c|}{15.4/11.9/7.0/ 13.6/16.5/17.9/17.7} \\
\bottomrule

\end{tabular}
\caption{Comparative data statistics for the shared Russian and English probing tasks. Label distribution by target class is presented in \%.}
\label{tab:data_description}
\end{table*}

\clearpage

\subsection{Probing Trajectories}
\begin{figure*}[h]
  \centering
  \includegraphics[width=.85\textwidth]{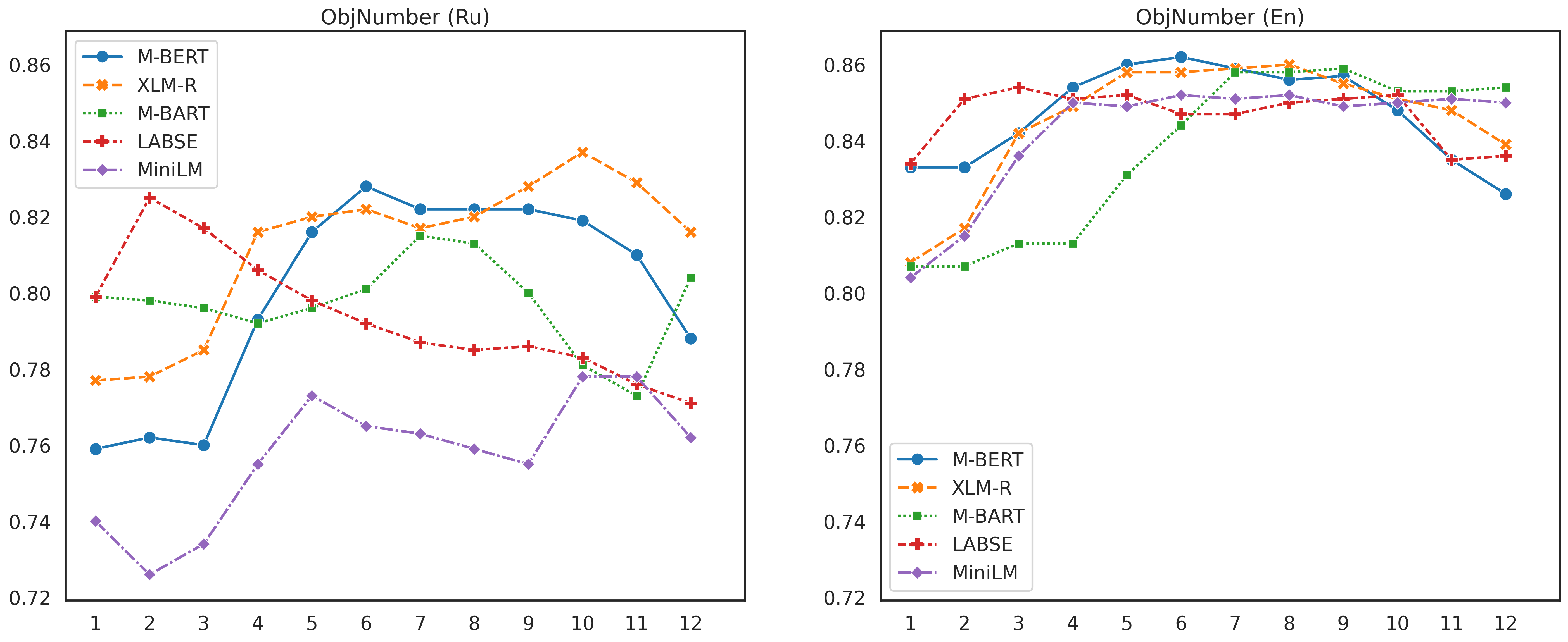}
  \caption{The probing results of Logistic Regression classifier for each encoder on \textbf{ObjNumber}. \textbf{Ru} is at the left; and \textbf{En} is at the right. X-axis=Layer number, Y-axis=Accuracy score.}
  \label{fig:ru_en_obj_number_logreg}

\vspace*{1 cm}
  \centering
  \includegraphics[width=.85\textwidth]{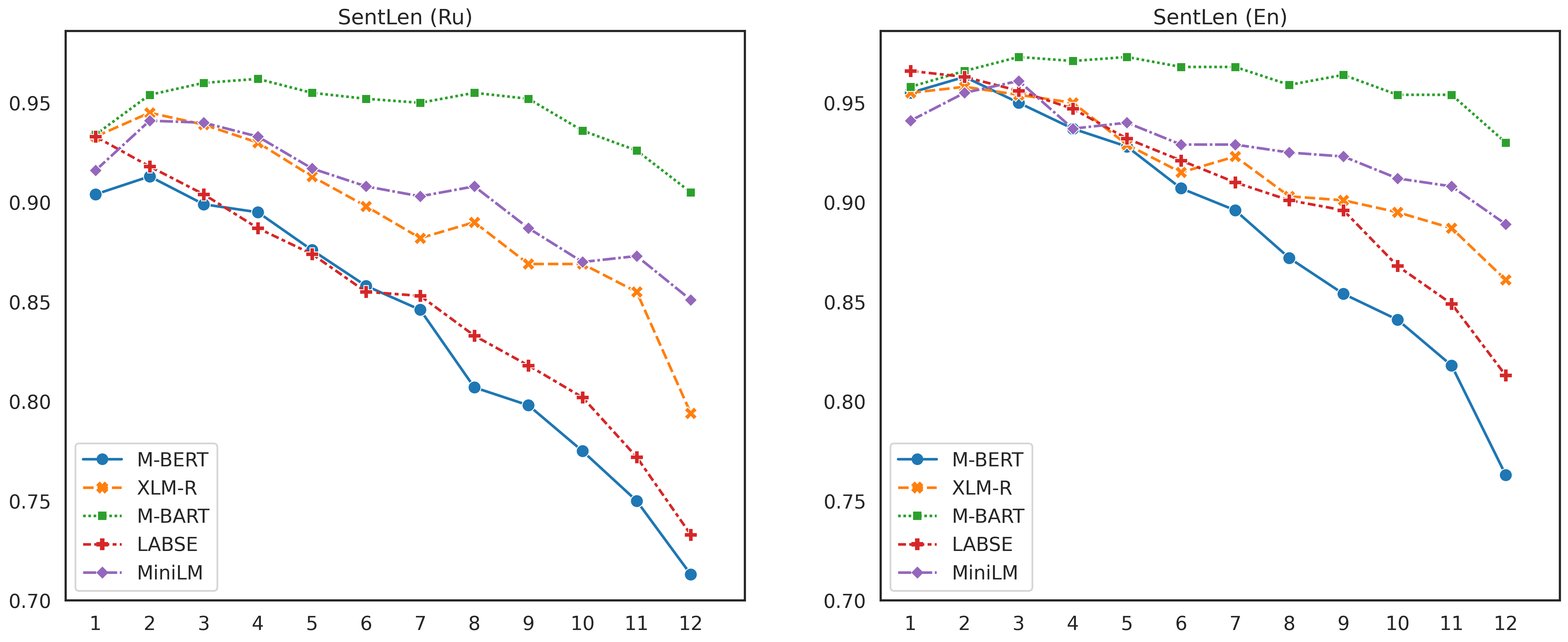}
  \caption{The probing results of Logistic Regression classifier for each encoder on \textbf{SentLen}. \textbf{Ru} is at the left; and \textbf{En} is at the right. X-axis=Layer number, Y-axis=Accuracy score.}
  \label{fig:ru_en_sent_len_logreg}
\end{figure*}

\begin{figure*}
  \centering
  \includegraphics[width=.85\textwidth]{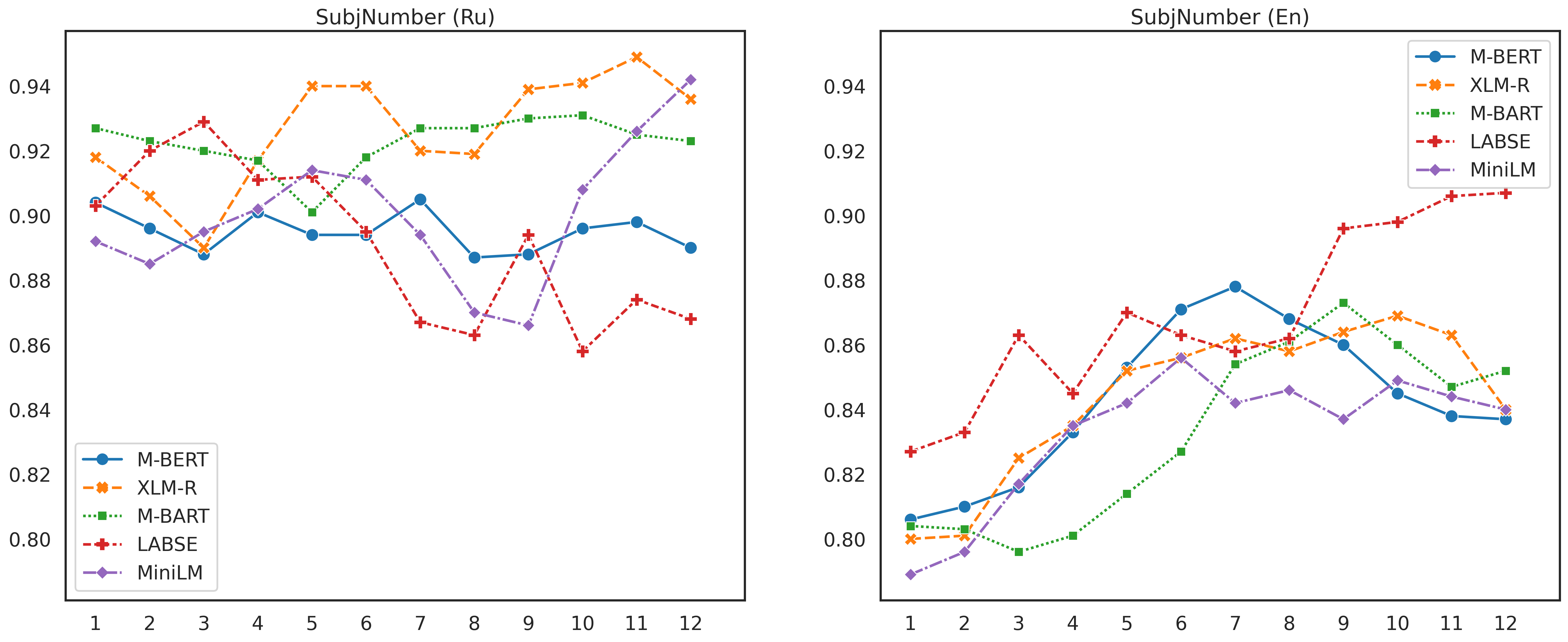}
  \caption{The probing results of Logistic Regression classifier for each encoder on \textbf{SubjNumber}. \textbf{Ru} is at the left; and \textbf{En} is at the right. X-axis=Layer number, Y-axis=Accuracy score.}
  \label{fig:ru_en_subj_number_logreg}
\vspace*{1 cm}
  \centering
  \includegraphics[width=.85\textwidth]{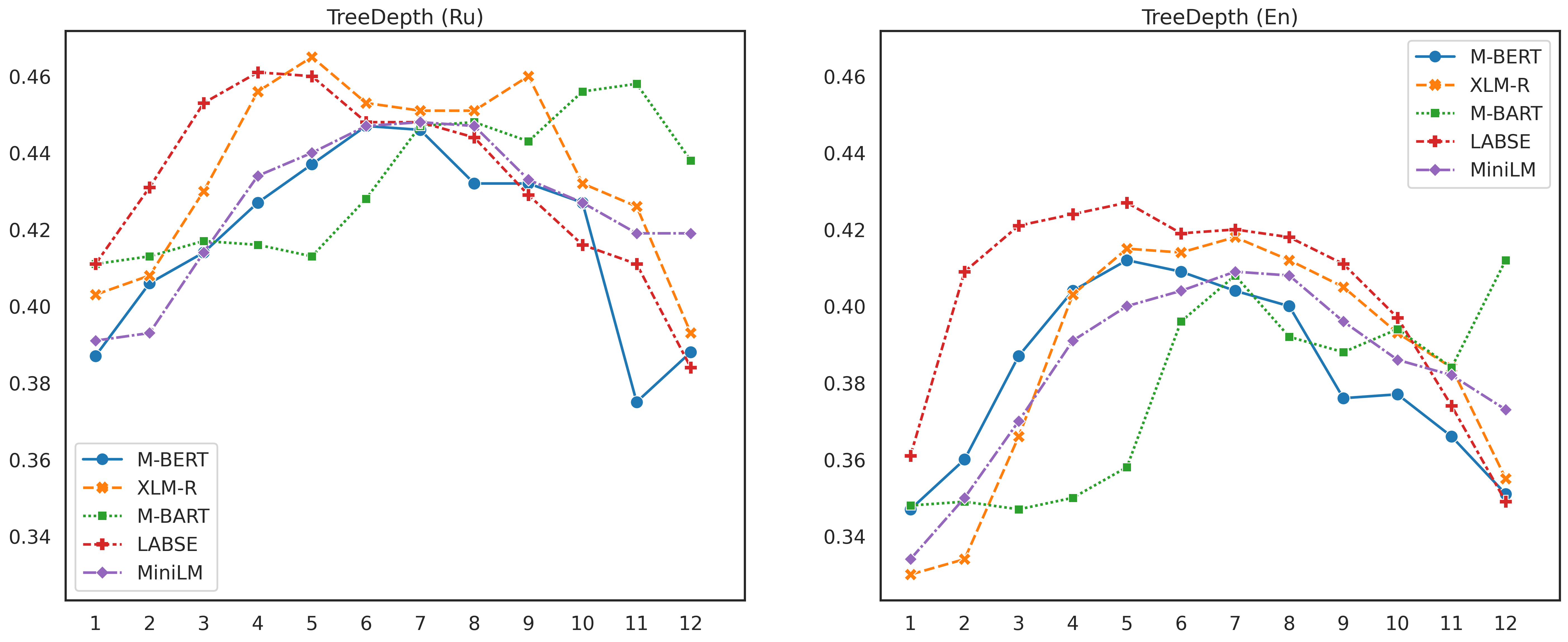}
  \caption{The probing results of Logistic Regression classifier for each encoder on \textbf{TreeDepth}. \textbf{Ru} is at the left; and \textbf{En} is at the right. X-axis=Layer number, Y-axis=Accuracy score.}
  \label{fig:ru_en_tree_depth_logreg}
\end{figure*}

\begin{figure*}
  \centering
  \includegraphics[width=.85\textwidth]{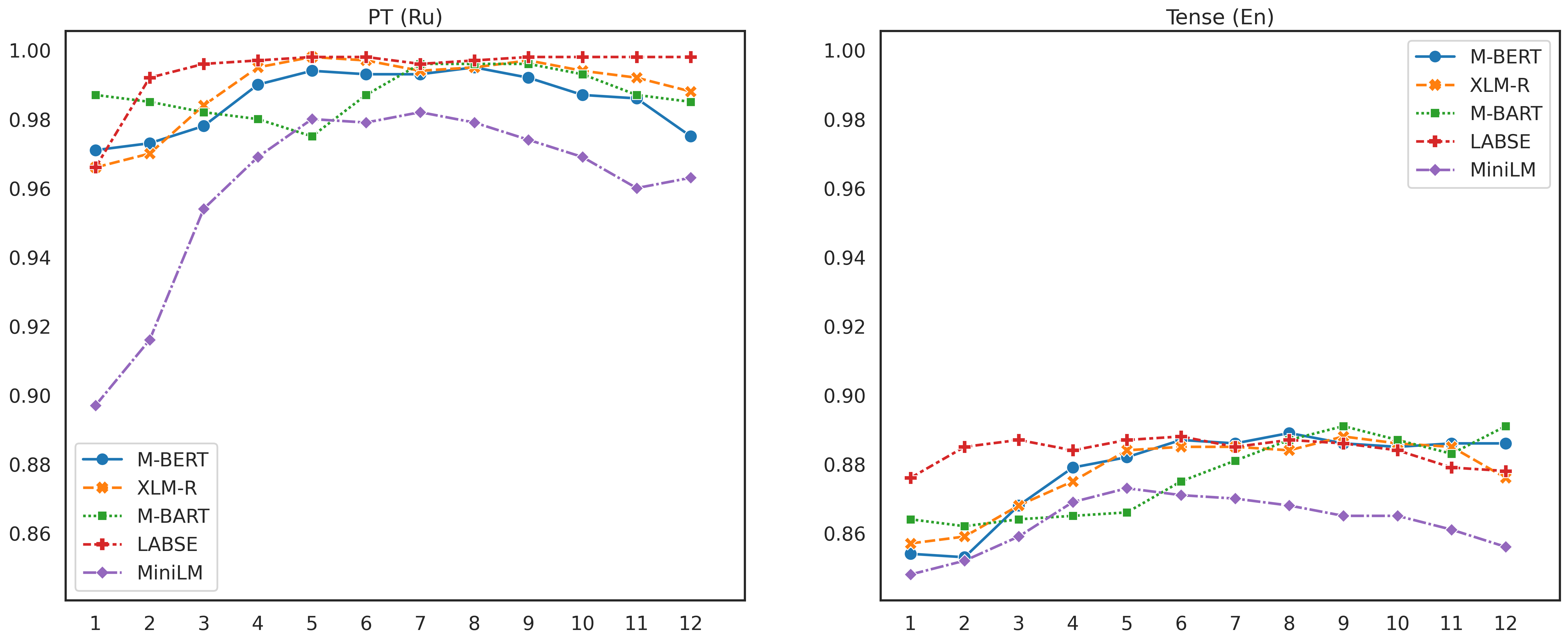}
  \caption{The probing results of Logistic Regression classifier for each encoder on \textbf{Tense}. \textbf{Ru} (\textbf{PT}) is at the left; and \textbf{En} (\textbf{Tense}) is at the right. X-axis=Layer number, Y-axis=Accuracy score.}
  \label{fig:ru_en_tense_logreg}
\vspace*{1 cm}
  \centering
  \includegraphics[width=.85\textwidth]{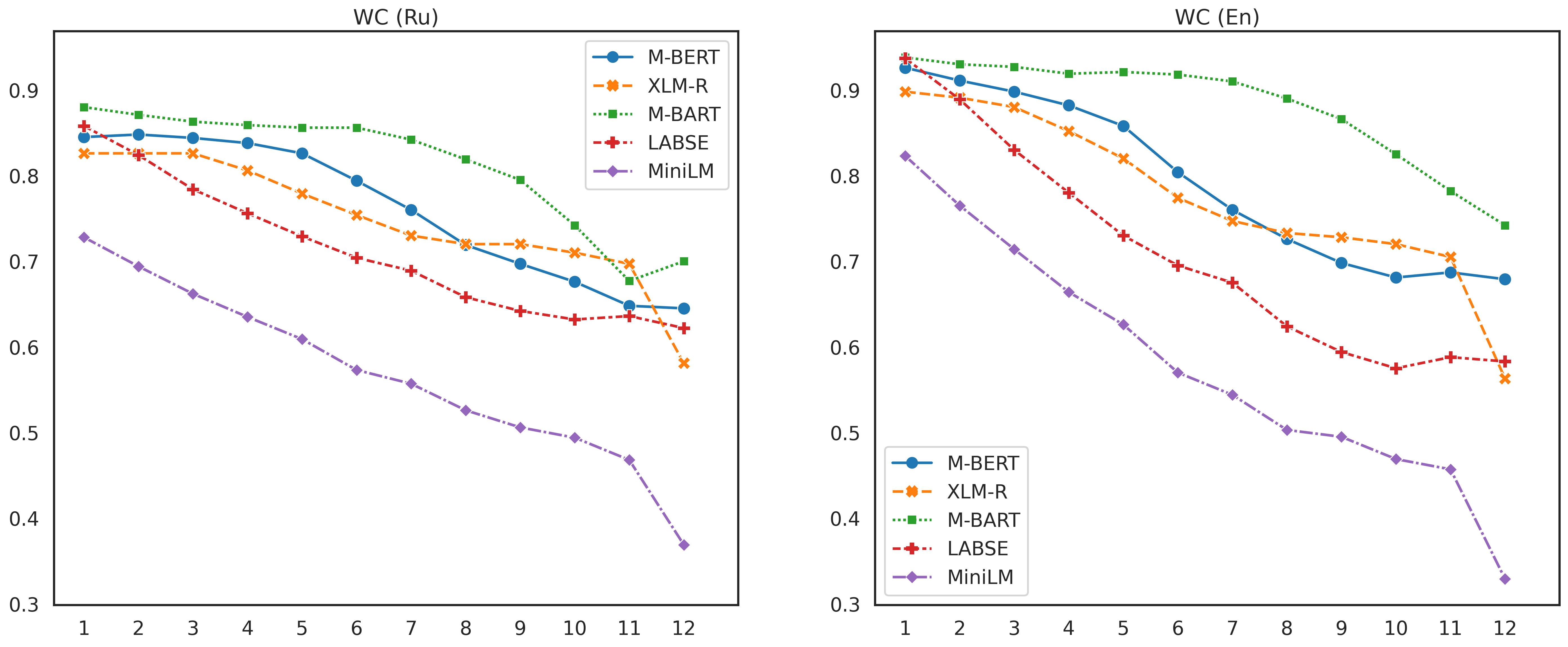}
  \caption{The probing results of Logistic Regression classifier for each encoder on \textbf{WC}. \textbf{Ru} is at the left; and \textbf{En} is at the right. X-axis=Layer number, Y-axis=Accuracy score.}
  \label{fig:ru_en_wc_logreg}
\end{figure*}

Figures \ref{fig:ru_en_obj_number_logreg}--\ref{fig:ru_en_wc_logreg} show the probing curves of Logistic Regression classifier over the shared tasks (see Section \ref{subsection:sup_probe}). The results obtained with MLP classifier are pretty consistent with the ones presented in this Appendix.

\clearpage
\section{Individual Neuron Analysis}
\label{appendix:neuron_analysis}
\begin{figure*}[h]
  \centering
  \includegraphics[width=\textwidth]{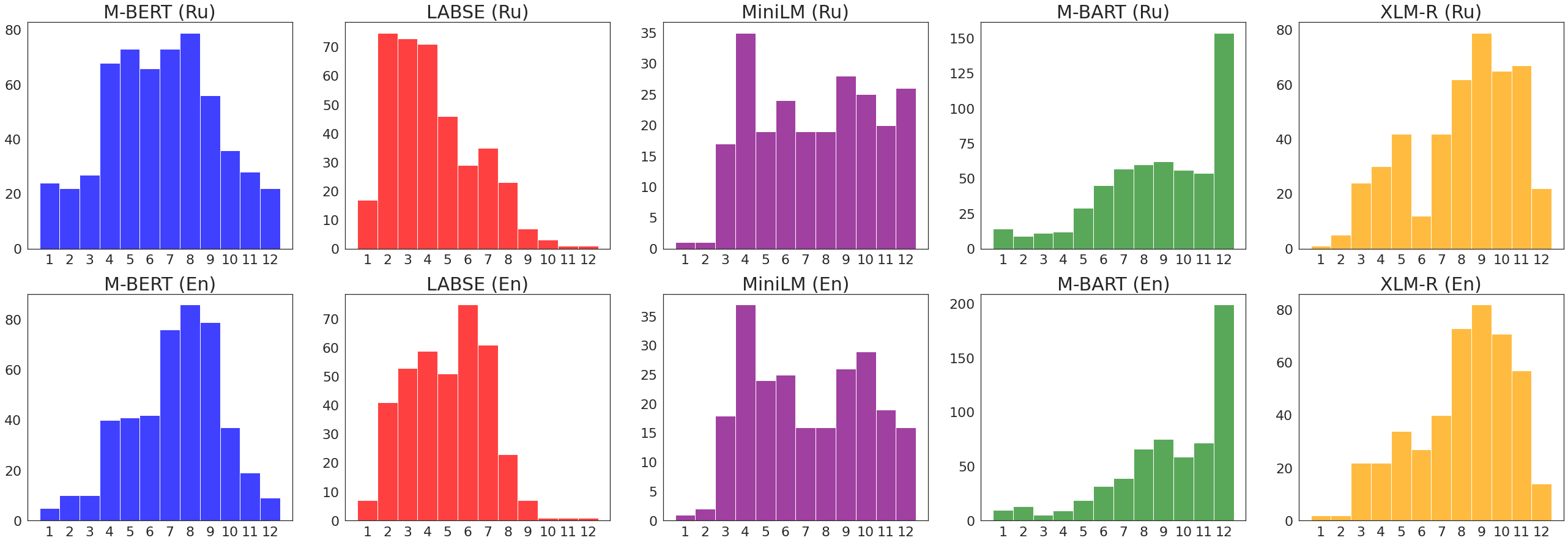}
  \caption{The distribution of top neurons over \textbf{NShift} tasks for both languages: \textbf{Ru}=Russian, \textbf{En}=English. X-axis=Layer index number, Y-axis=Number of neurons selected from the layer.}
  \label{fig:neuron_ngram_shift}
\vspace*{1 cm}
  \centering
  \includegraphics[width=\textwidth]{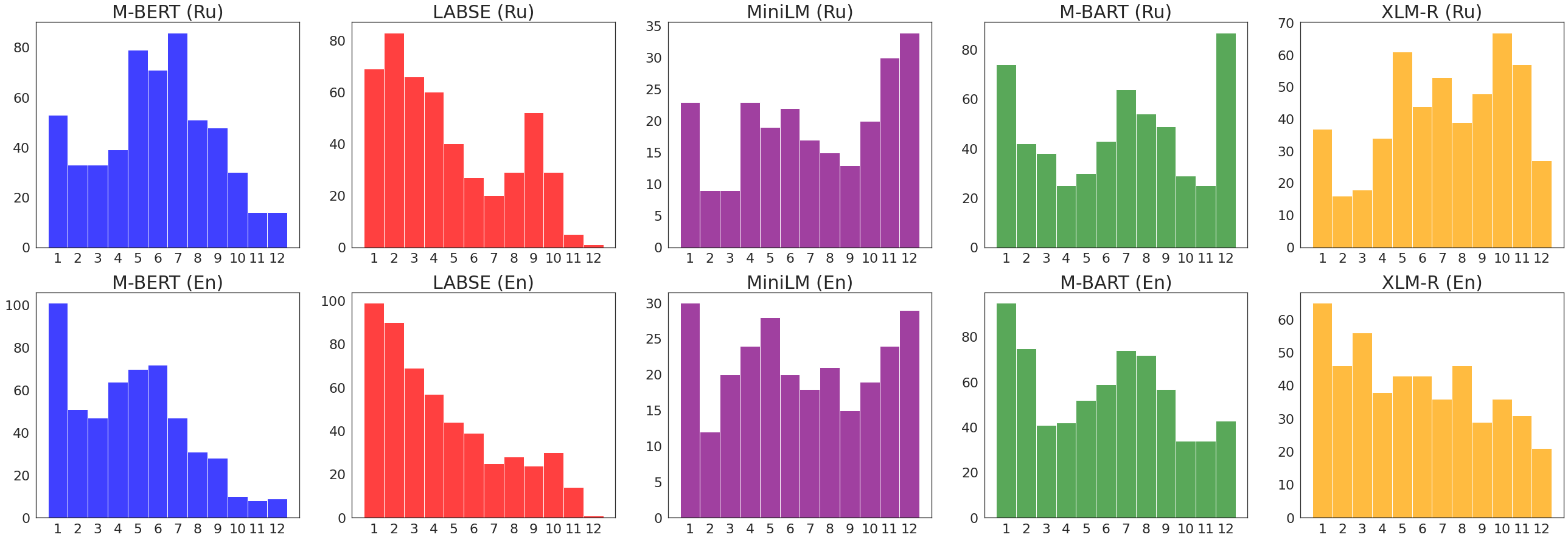}
  \caption{The distribution of top neurons over \textbf{ObjNumber} tasks for both languages: \textbf{Ru}=Russian, \textbf{En}=English. X-axis=Layer index number, Y-axis=Number of neurons selected from the layer.}
  \label{fig:neuron_obj_number}
\end{figure*}

\begin{figure*}
  \centering
  \includegraphics[width=\textwidth]{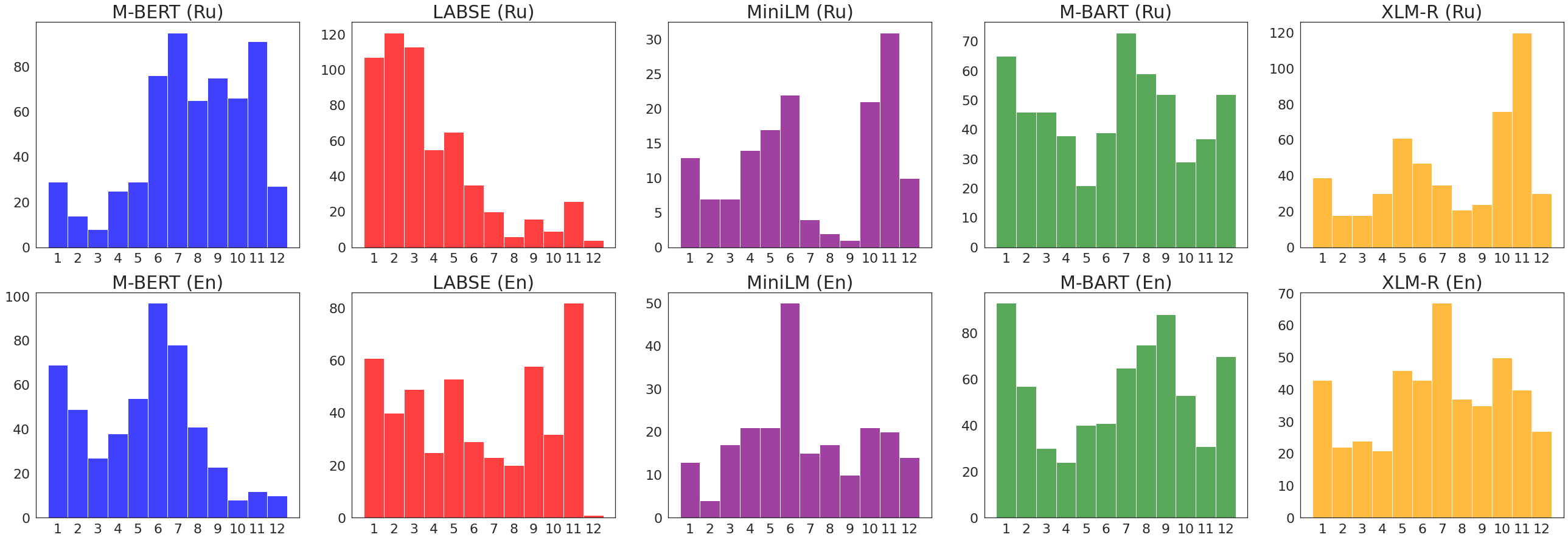}
  \caption{The distribution of top neurons over \textbf{SubjNumber} tasks for both languages: \textbf{Ru}=Russian, \textbf{En}=English. X-axis=Layer index number, Y-axis=Number of neurons selected from the layer.}
  \label{fig:neuron_subj_number}
\vspace*{1 cm}
  \centering
  \includegraphics[width=\textwidth]{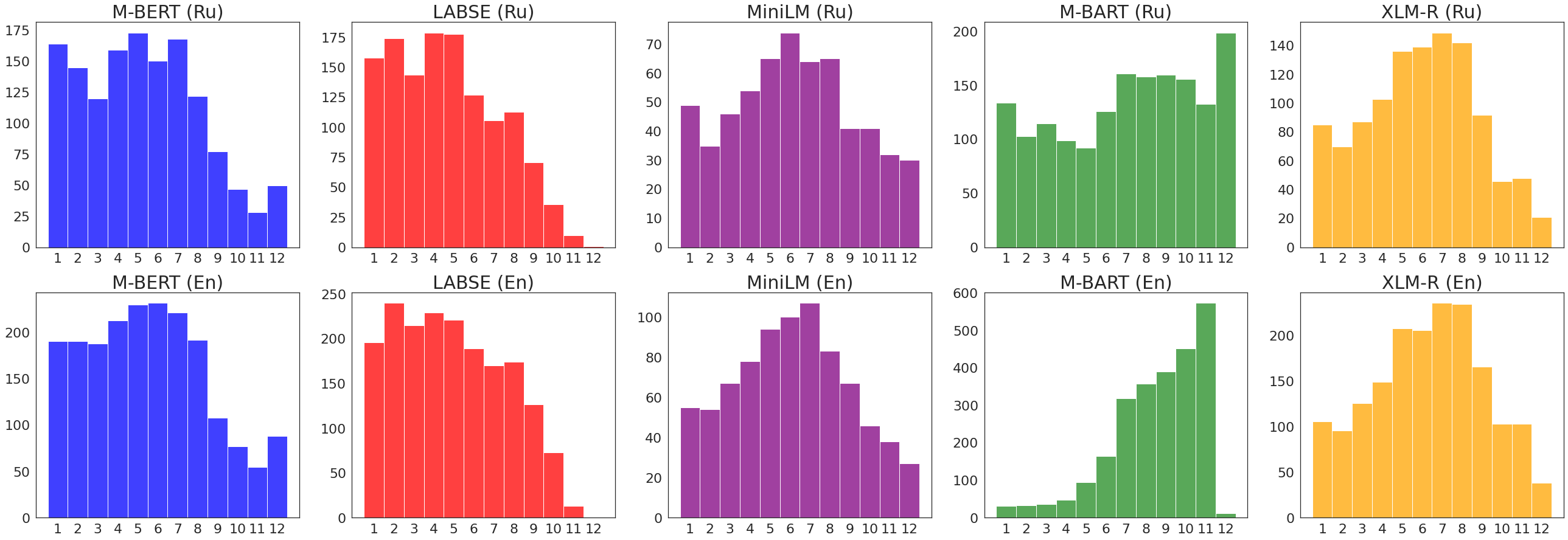}
  \caption{The distribution of top neurons over \textbf{TreeDepth} tasks for both languages: \textbf{Ru}=Russian, \textbf{En}=English. X-axis=Layer index number, Y-axis=Number of neurons selected from the layer.}
  \label{fig:neuron_tree_depth}
\vspace*{1 cm}
  \centering
  \includegraphics[width=\textwidth]{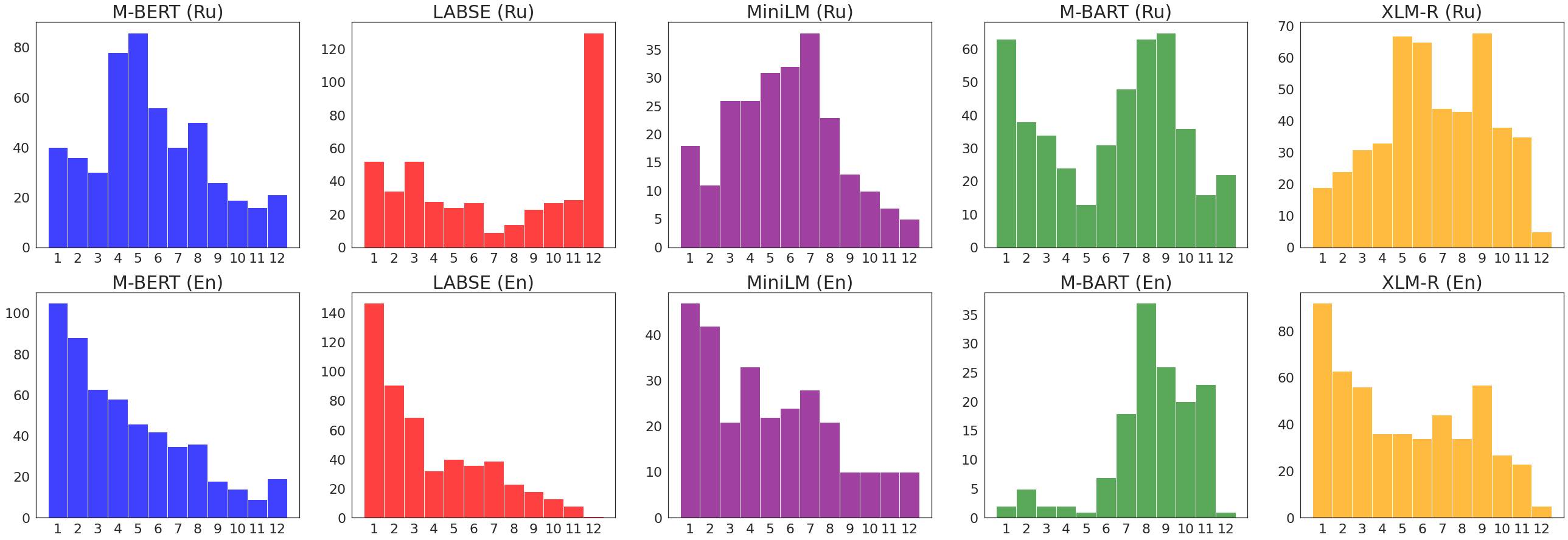}
  \caption{The distribution of top neurons over \textbf{Tense} tasks for both languages: \textbf{Ru}=Russian, \textbf{En}=English. X-axis=Layer index number, Y-axis=Number of neurons selected from the layer.}
  \label{fig:neuron_tense}
\end{figure*}

Figures \ref{fig:neuron_ngram_shift}--\ref{fig:neuron_tense} depict top neuron distributions for the tasks (see Section \ref{subsec:neuron_analysis}).

\clearpage

\section{Correlation Methods}
\label{appendix:corr_methods}
\begin{figure*}[h]
    \centering
    \begin{subfigure}[b]{0.49\linewidth}
    \centering
    \includegraphics[width=\linewidth]{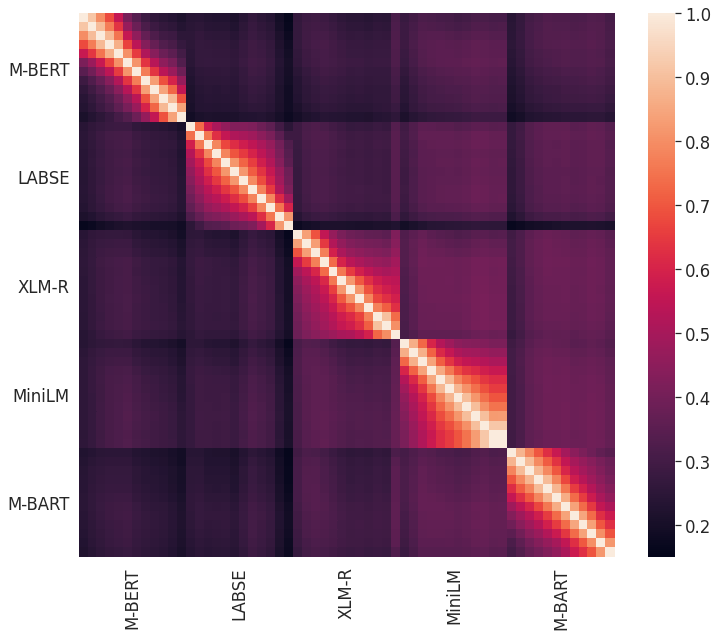}
    \caption{\texttt{maxcorr}}
    \label{fig:heatmap-maxcorr-ru}
    \end{subfigure}
    \begin{subfigure}[b]{0.49\linewidth}
    \centering
    \includegraphics[width=\linewidth]{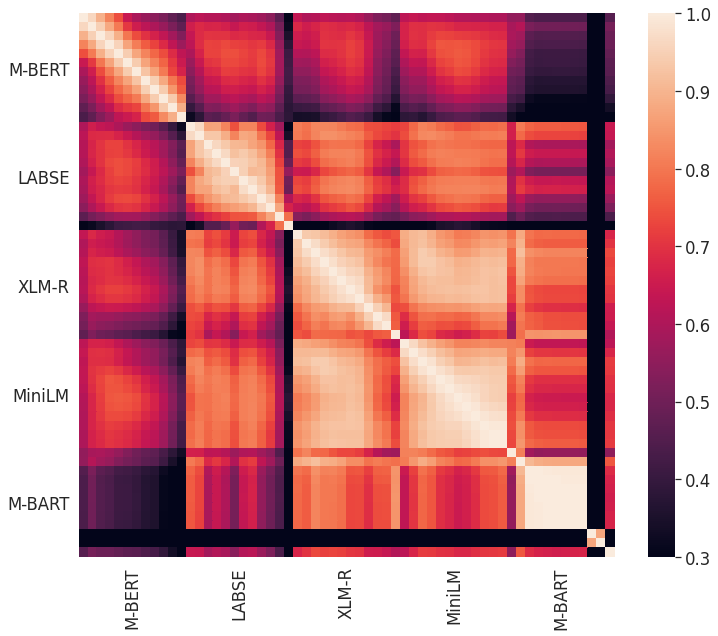}
    \caption{\texttt{lincka}}
    \label{fig:heatmap-lincka-ru}
    \end{subfigure}    
    \caption{Similarity heatmaps of layers in the encoders under neuron-level (\texttt{maxcorr})
    and representation-level (\texttt{lincka}) measures for Russian.
    }
    \label{fig:heatmaps-ru}
\end{figure*}

Heatmaps (Figure \ref{fig:heatmaps-ru}) show similarities of the encoders under neuron-level and representation-level correlation-based similarity measures on the Russian tasks (see Section \ref{subsec:corr_methods}).

\end{document}